\pdfoutput=1
\documentclass[11pt]{article}

\usepackage[preprint]{acl}

\usepackage{times}
\usepackage{latexsym}
\usepackage[T1]{fontenc}
\usepackage[utf8]{inputenc}
\usepackage{microtype}
\usepackage{inconsolata}
\usepackage{graphicx}

\usepackage{multirow}
\usepackage{amsmath}
\usepackage{dsfont}
\usepackage{booktabs}
\usepackage{amssymb}
\usepackage{makecell}
\usepackage{tabularx}
\usepackage{subcaption}
\usepackage{bm}
\usepackage{tcolorbox}
\usepackage{ragged2e}

\definecolor{lightgray}{rgb}{0.9,0.9,0.9}

\newcommand{\norm}[1]{\left\lVert#1\right\rVert}

\newcommand*\diff{\mathop{}\!\mathrm{d}}

\newcommand{\NA}{\rule[0.5ex]{0.7em}{0.5pt}}

\title{3D-MoE: Towards Spatial Intelligence with Mixture-of-Experts for 3D Reasoning and Action Generation}

\author{
Yueen Ma\textsuperscript{\rm 1,\rm 3},
Zenglin Xu\textsuperscript{\rm 2,\rm 3},
Irwin King\textsuperscript{\rm 1}\\
\textsuperscript{\rm 1}The Chinese University of Hong Kong\\
\textsuperscript{\rm 2}Fudan University\\
\textsuperscript{\rm 3}Shanghai Academy of AI for Science\\
{\normalfont\normalsize
\texttt{mayueen@link.cuhk.edu.hk},\quad
\texttt{zenglinxu@fudan.edu.cn},\quad
\texttt{king@cse.cuhk.edu.hk}}
}

\begin{document}
\maketitle
\begin{abstract}
Spatial intelligence, encompassing 3D perception and reasoning, is the essential next frontier of AI. Scaling current 3D vision-language models (VLMs) that rely on dense Transformers for spatial tasks incurs prohibitive computational costs. In this paper, we introduce 3D-MoE, a 3D VLM leveraging an efficient mixture-of-experts architecture with a modality- and spatial-context-aware probabilistic routing scheme, stably cultivated by a novel routing curriculum. To seamlessly extend 3D-MoE to embodied AI, we integrate a diffusion-based action head, Pose-DiT, transforming 3D-MoE into a 3D vision-language-action (VLA) model. By employing a rectified flow framework, Pose-DiT generates precise 6D pose actions in a single sampling step. Extensive experiments demonstrate that 3D-MoE achieves superior performance on diverse 3D vision-language benchmarks with drastically fewer activated parameters and yields higher success rates while enabling real-time inference for robot manipulation tasks. Code is available at \url{https://github.com/yueen-ma/3D-MoE}.
\end{abstract}

\section{Introduction}
\label{sec:intro}

\begin{figure}[t]
 \centering
    \includegraphics[width=1.0\columnwidth, trim={0.435in 7in 4.935in 1in},clip]{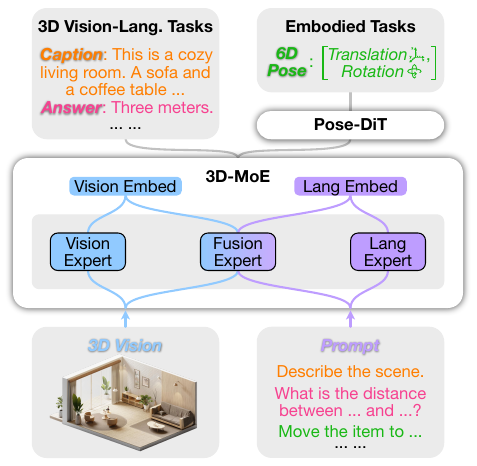}
    \caption{3D-MoE accommodates 3D visual inputs and either generates textual responses for various 3D vision-language tasks---such as 3D scene captioning and 3D question answering---or predicts 6D poses for embodied tasks using the Pose-DiT action diffusion module. ``Vision Embed'' and ``Lang Embed'' refer to vision (blue) and language (purple) embeddings, respectively.
    }
    \label{fig:intro}
\end{figure}

\begin{figure*}
 \centering
    \includegraphics[width=1.0\textwidth, trim={0.4in 5.35in 2.75in 0in},clip]{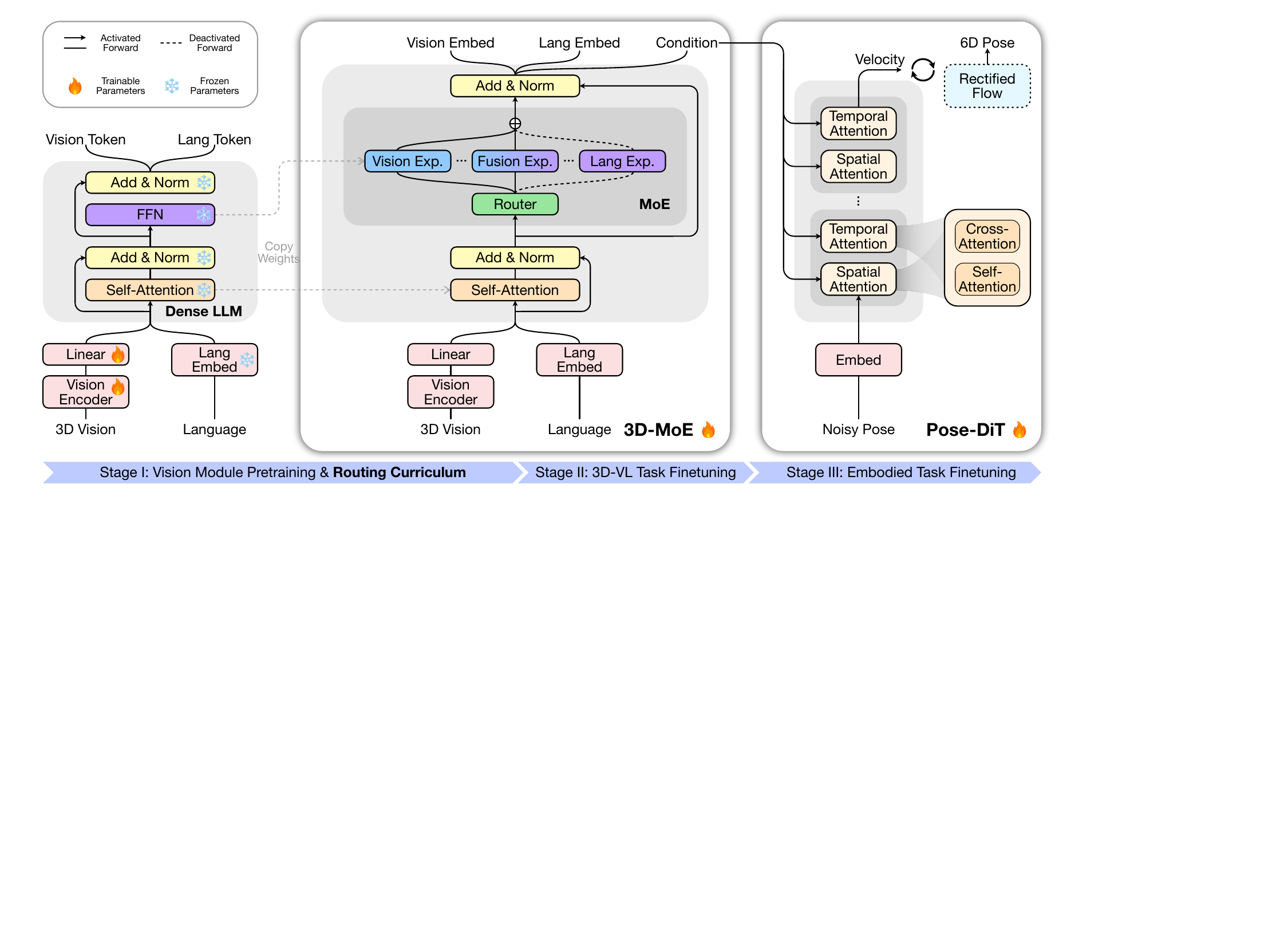}
    \caption{\textbf{Architecture of 3D-MoE and routing curriculum in a three-stage training pipeline.} In Stage I, we pretrain the vision modules and apply our proposed routing curriculum to warm-start 3D-MoE. In Stage II, we finetune 3D-MoE on 3D vision-language tasks. In Stage III, we integrate the Pose-DiT action head, enabling 3D-MoE to predict 6D pose actions for embodied tasks. $\oplus$ denotes a weighted average.
    }
    \label{fig:overview}
\end{figure*}

Spatial intelligence \cite{DBLP:journals/corr/abs-2412-14171, world-labs-about} represents the next frontier of AI, encompassing capabilities beyond the verbal reasoning of large language models (LLMs), such as 3D visual perception, spatial reasoning, and visuospatial common sense. These capabilities are essential for solving complex 3D vision-language and embodied tasks. While existing vision-language models (VLMs) \cite{DBLP:journals/corr/abs-2304-08485, DBLP:journals/corr/abs-2412-15115} exhibit some spatial capabilities, their reliance on 2D images fundamentally limits their potential. Because images are lossy 2D projections of a 3D world, they strip away critical geometric information, such as precise 3D object locations, thereby restricting a model's performance in spatial reasoning and physical manipulation.

To overcome these limitations, recent works have introduced 3D VLMs \cite{DBLP:conf/icmcs/WangLXQYMLZ25, DBLP:conf/cvpr/DengHJWDR25} that directly integrate 3D vision encoders with dense LLMs. Although these models achieve strong performance on 3D vision-language (3D-VL) tasks like 3D question answering and scene captioning, they inherit the prohibitive computational costs associated with scaling dense Transformer architectures.

To efficiently scale model capacity, researchers have increasingly adopted the Mixture-of-Experts (MoE) architecture \cite{DBLP:journals/corr/abs-2401-04088}. By sparsely activating only a top-$k$ subset of experts, MoE significantly reduces inference compute. Furthermore, MoE is exceptionally well-suited for multimodal learning \cite{DBLP:conf/cvpr/LuoYDWLD0Z25, DBLP:journals/corr/abs-2511-12609, DBLP:journals/corr/abs-2412-10302} because it mitigates modality interference by routing distinct modalities to dedicated experts, avoiding the capacity competition inherent to shared FFN layers. Current multimodal MoEs typically employ one of two routing paradigms. Deterministic routing rigidly partitions networks into modality-specific or fusion experts, ensuring training stability but limiting adaptability. Conversely, probabilistic routing allows the router to automatically learn token assignments, offering flexibility but risking expert collapse \cite{DBLP:journals/corr/abs-2507-12566, DBLP:conf/nips/MustafaRPJH22}.

In this paper, we propose unifying the stability of deterministic routing with the flexibility of probabilistic routing through a novel \textit{routing curriculum}. This curriculum begins with a \textit{specialist warmup} phase, utilizing deterministic routing to stably cultivate experts strictly specialized for vision or language via a hard partition boundary that separates these two groups of experts. This is followed by a \textit{differential boundary relaxation} phase, which gradually transitions into probabilistic routing by lowering this partition boundary. This relaxation cultivates fusion experts capable of handling multimodal representations, while preserving a subset of modality-specific specialists. Viewed from another perspective, this phase gradually removes the potentially suboptimal human priors about expert partitioning imposed during the warmup phase. Complemented by penalty-adaptive auxiliary-loss-free balancing and explicit spatial routing signals, this curriculum yields \textbf{3D-MoE}: an efficient architecture with modality- and spatial-context-aware routing that excels at 3D vision-language tasks.

Because spatial intelligence is a cornerstone of embodied AI, we also transfer the 3D-VL knowledge captured by 3D-MoE to address complex embodied tasks. Unlike previous VLAs \cite{DBLP:journals/corr/abs-2307-15818, DBLP:journals/corr/abs-2406-09246} that predominantly rely on 2D inputs, 3D-MoE natively processes 3D observations, providing the precise geometric perception required for robotic manipulation. To realize a VLA architecture, we equip the 3D-MoE backbone with \textbf{Pose-DiT}, a diffusion-based action prediction module. While recent VLAs have explored flow-matching paradigms \cite{DBLP:journals/corr/abs-2410-24164}, they often require multiple sampling steps. We adapt rectified flow \cite{DBLP:conf/iclr/LiuG023}, previously used to speed up video generation \cite{opensora}, to enable single-step 6D pose action prediction, drastically improving inference speed to achieve real-time execution critical for embodied control. Furthermore, inspired by ST-DiT \cite{DBLP:journals/corr/abs-2412-20404}, Pose-DiT utilizes explicit spatial and temporal Transformer layers to ensure enhanced spatiotemporal consistency in action sequences.

The main contributions of this paper are:
\begin{itemize}
    \item We introduce 3D-MoE, an MoE-based MLLM designed for spatial tasks. It features modality- and spatial-context-aware routing cultivated via a novel routing curriculum that ensures both stability and flexibility.
    \item We propose Pose-DiT, a spatiotemporal diffusion action head that extends 3D-MoE into a 3D VLA. By leveraging the rectified flow framework, Pose-DiT achieves high-quality action generation with single-step inference.
    \item We demonstrate the effectiveness and efficiency of our approach through extensive experiments on diverse 3D vision-language benchmarks and complex embodied tasks.
\end{itemize}

\section{Related Work}

\paragraph{3D VLMs.} Recent progress in spatial intelligence \cite{DBLP:journals/corr/abs-2412-14171, world-labs-about} has led to multimodal LLMs integrated with 3D vision \cite{DBLP:conf/icml/HuangYMLLW0ZJ024, DBLP:conf/nips/HuangC0HXWLCZPZ24, DBLP:conf/icmcs/WangLXQYMLZ25, DBLP:conf/cvpr/DengHJWDR25}. In this work, we build upon established 3D vision encoders, focusing instead on optimizing backbone efficiency via the MoE architecture.

\paragraph{MoE.} Mixture-of-Experts architectures mitigate the scaling costs of dense Transformers by sparsely activating parameters without sacrificing performance. Pioneering works like Flan-MoE \cite{DBLP:journals/corr/abs-2305-14705}, OpenMoE \cite{DBLP:conf/icml/XueZFNZZ024}, and Mixtral \cite{DBLP:journals/corr/abs-2401-04088} have been recently followed by DeepSeek-V3 \cite{DBLP:journals/corr/abs-2412-19437}, the MoE variants of Qwen2.5 \cite{DBLP:journals/corr/abs-2412-15115}, and Llama 4 \cite{llama4-meta}. In multimodal settings, MoE architectures mitigate modality interference by routing distinct data types to dedicated experts, thereby avoiding the capacity competition inherent to shared FFN layers. Existing multimodal MoE models employ either deterministic routing (e.g., Mono-InternVL \cite{DBLP:conf/cvpr/LuoYDWLD0Z25}, BEiT-3 \cite{DBLP:conf/cvpr/WangBDBPLAMSSW23}) or probabilistic routing (e.g., DeepSeek-VL2 \cite{DBLP:journals/corr/abs-2412-10302}, Uni-MoE-2.0-Omni \cite{DBLP:journals/corr/abs-2511-12609}). We unify these paradigms under a novel routing curriculum to combine their advantages.

\paragraph{VLA.} Vision-Language-Action models \cite{DBLP:journals/tnn/MaSZHK26} extend VLMs to generate robot actions. RT-2 \cite{DBLP:journals/corr/abs-2307-15818} pioneered this approach and was followed by models like OpenVLA \cite{DBLP:journals/corr/abs-2406-09246}, $\pi_0$ \cite{DBLP:journals/corr/abs-2410-24164}, and Astra \cite{ma-etal-2025-astra}. Building upon existing VLMs, they predominantly rely on 2D visual inputs; in contrast, 3D-MoE natively accommodates 3D vision. For action generation, diffusion-based policies emerged in embodied AI \cite{DBLP:conf/rss/ChiFDXCBS23, Ze2024DP3, DBLP:journals/corr/abs-2402-10885} following their success in computer vision (CV). Adapting DiTs \cite{DBLP:conf/iccv/PeeblesX23} for robot learning, models like RDT \cite{liu2024rdt} and MDT \cite{DBLP:journals/corr/abs-2407-05996} rely on classical DDPM \cite{DBLP:conf/nips/HoJA20} or DDIM \cite{DBLP:conf/iclr/SongME21} schedulers, which struggle to achieve the desired inference quality and speed. While $\pi_0$ adopts flow matching, it still requires multiple sampling steps to predict actions. Conversely, our Pose-DiT leverages rectified flow \cite{DBLP:conf/iclr/LiuG023}, requiring only a single sampling step.

\begin{figure*}
    \centering
    \begin{subfigure}[t]{\textwidth}
        \centering
        \includegraphics[width=\textwidth, trim={0.45in 8.0in 3in 0.75in},clip]{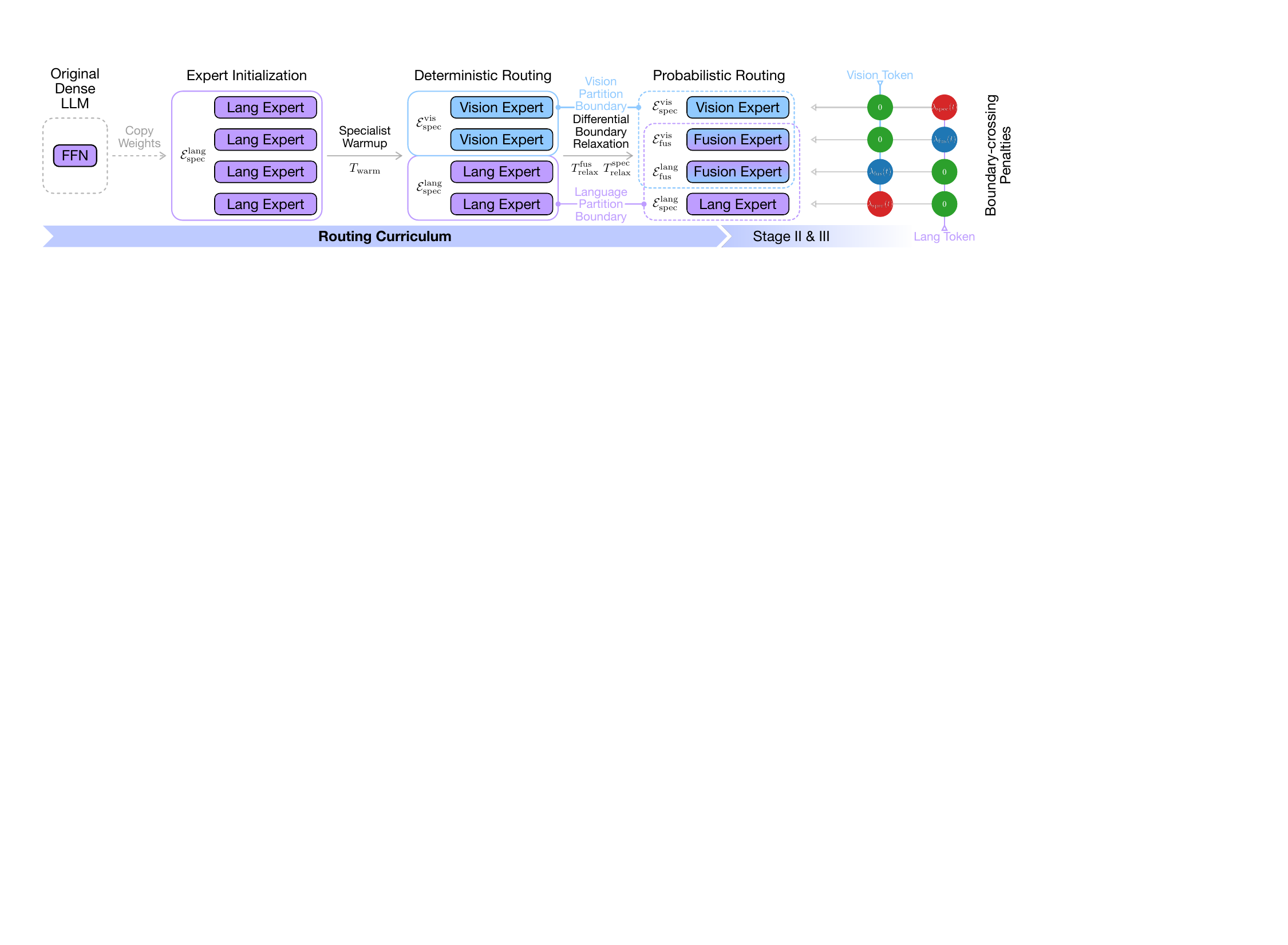}
        \caption{}
        \label{fig:experts}
    \end{subfigure}

    \par\vspace{0.2cm}

    \begin{subfigure}[t]{\textwidth}
        \centering
        \includegraphics[width=\textwidth]{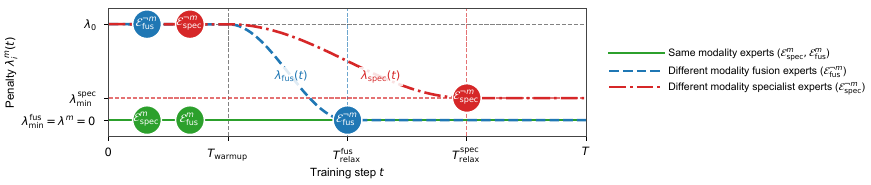}
        \caption{}
        \label{fig:cosine}
    \end{subfigure}
    \caption{\textbf{Routing curriculum of 3D-MoE.} (a) Overview of the routing curriculum for progressively guiding the development of modality- and spatial-context-aware routing across two groups of specialist experts ($\mathcal{E}_{\text{spec}}^{\text{vis}}$ and $\mathcal{E}_{\text{spec}}^{\text{lang}}$) and two groups of fusion experts ($\mathcal{E}_{\text{fus}}^{\text{vis}}$ and $\mathcal{E}_{\text{fus}}^{\text{lang}}$). (b) Cosine decay schedule for the boundary-crossing penalties $\lambda_i^m(t)$, which is defined by the following key parameters: $\lambda_0$, $T_{\text{warm}}$, $\lambda^{\text{fus}}_{\min}$, $T^{\text{fus}}_{\text{relax}}$, $\lambda^{\text{spec}}_{\min}$, $T^{\text{spec}}_{\text{relax}}$.}
    \label{fig:curriculum}
\end{figure*}

\section{Method}

\subsection{Overview}
3D-MoE consists of three main components: a 3D vision module, an MoE LLM backbone, and a diffusion module, as shown in Figure~\ref{fig:overview}. By integrating the vision encoder with the MoE backbone, 3D-MoE is capable of handling a variety of 3D vision-language (3D-VL) tasks. When equipped with our Pose-DiT action diffusion module, 3D-MoE can also address embodied tasks.

\subsection{Architecture of the 3D-MoE Backbone}

To efficiently scale 3D multimodal learning, we construct 3D-MoE by initializing it with a pretrained dense LLM. We expand model capacity while preserving established linguistic knowledge by replicating the dense feed-forward network (FFN) weights across multiple expert FFNs \cite{DBLP:journals/corr/abs-2401-15947, DBLP:journals/corr/abs-2405-11273}. Following LLaVA \cite{DBLP:journals/corr/abs-2304-08485} and LEO \cite{DBLP:conf/icml/HuangYMLLW0ZJ024}, a single linear projection connects the 3D vision encoder to the LLM.

For a token embedding $\mathbf{x} \in \mathbb{R}^d$, our MoE layer sparsely activates a subset $\mathcal{K}$ containing the indices of the $k$ experts with the highest routing logits:
\begin{equation}
\text{MoE}(\mathbf{x}) = \sum_{i \in \mathcal{K}} \text{softmax}_i(\mathbf{z}(\mathbf{x}, t)) \cdot \mathcal{E}_i(\mathbf{x}),
\end{equation}
where $\mathcal{E}_i(\mathbf{x})$ denotes the output of the $i$-th expert, and $\mathbf{z}(\mathbf{x}, t) \in \mathbb{R}^E$ is the vector of routing logits at training step $t$.

\subsection{Routing Curriculum}

Experts initialized from the LLM's FFN layers inherently possess a strong linguistic bias because the original FFNs processed only language tokens. Abruptly sending vision tokens to a randomly initialized MoE router may cause expert collapse \cite{DBLP:journals/corr/abs-2507-12566, DBLP:conf/nips/MustafaRPJH22}, wherein the router activates only a fixed subset of the experts while ignoring others. To address this issue, we introduce a routing curriculum that progressively guides 3D-MoE to develop a modality- and spatial-context-aware probabilistic routing scheme.

The routing curriculum involves two main phases: specialist warmup and differential boundary relaxation, as illustrated in Figure~\ref{fig:experts}. A partition boundary for a token of modality $m \in \{\text{lang}, \text{vis}\}$ restricts the subset of experts to which the token may be routed. Accordingly, the curriculum is governed by adjusting a boundary-crossing penalty $\lambda_i^m(t)$ throughout these two phases. Further incorporating spatial routing signals $g(\mathbf{x}) \in \mathbb{R}^S$ and a penalty-adaptive balancing bias $b_i(t)$, we formulate the overall routing logit for expert $i$ as:
\begin{equation}
z_i(\mathbf{x}, t) = \mathbf{W}_{r, i}^\top \text{LN}(\mathbf{x} \parallel g(\mathbf{x})) - \lambda_i^m(t) + b_i(t),
\end{equation}
where $\mathbf{W}_r \in \mathbb{R}^{(d+S) \times E}$ is the learnable routing projection for $E$ experts, with $\mathbf{W}_{r,i}$ as its $i$-th column; $\parallel$ denotes concatenation; and $\text{LN}(\cdot)$ represents layer normalization.

\paragraph{Specialist Warmup.}
The specialist warmup phase relieves each router of the burden of prematurely determining which experts are optimal for each token. Since all experts are initially specialized for language tokens, training a router to select optimal experts for vision tokens at this early stage is largely arbitrary and unstable. Instead, we use a predefined, deterministic routing scheme to cultivate experts specialized for vision tokens while preserving the LLM's linguistic capabilities within the language experts.

Specifically, we manually partition the experts into language- and vision-specialist experts (hereafter referred to simply as specialists), $\mathcal{E}_{\text{spec}} = \mathcal{E}_{\text{spec}}^{\text{lang}} \cup \mathcal{E}_{\text{spec}}^{\text{vis}}$, by setting a boundary around each specialist partition. During the warmup phase, which ends at training step $T_{\text{warm}}$, if a router attempts to assign a token of modality $m$ to an opposite-modality expert $i \in \mathcal{E}_{\text{spec}}^{\neg m}$, a large boundary-crossing penalty $\lambda_i^m(t) = \lambda_0 \gg 0$ is imposed. This effectively overrides the learnable weights of the respective router, $\mathbf{W}_r$, akin to deterministically routed MoE models like Mono-InternVL \cite{DBLP:conf/cvpr/LuoYDWLD0Z25}. To avoid routing tokens to heavily penalized experts, we ensure that $k \leq |\mathcal{E}_{\text{spec}}^m|$.

\paragraph{Differential Boundary Relaxation.}
Relying exclusively on specialists can hinder the processing of highly multimodal data. To address this, our approach draws inspiration from the ``shared experts'' employed in recent MoE-based VLMs (e.g., DeepSeek-VL2 \cite{DBLP:journals/corr/abs-2412-10302}, Uni-MoE-2.0-Omni \cite{DBLP:journals/corr/abs-2511-12609}). However, rather than using these always-active shared experts, we develop conditionally activated \textit{fusion experts}, $\mathcal{E}_{\text{fus}}$, dedicated to processing multimodal embeddings. These fusion experts are derived from both a subset of the language specialists and a subset of the vision specialists, denoted as $\mathcal{E}_{\text{fus}}^{\text{lang}}$ and $\mathcal{E}_{\text{fus}}^{\text{vis}}$, respectively. This strategy guides fusion experts from both modalities to ``meet in the multimodal midground'' from opposite directions, improving the diversity of the resulting fusion experts.

To create these fusion experts after the warmup phase, we expand the routing boundaries to expose these designated experts to tokens from both modalities. Furthermore, because the specialists were manually partitioned during warmup, the initial boundaries may be suboptimal. Therefore, we gradually allow the routers to override these predefined boundaries where necessary. This process effectively removes the human priors regarding expert partitioning that were imposed during the warmup phase. However, to ensure the remaining specialists retain their modality-specific expertise, the boundary relaxation applied to them is intentionally constrained, resulting in a \textit{differential boundary relaxation} strategy.

Specifically, for a token of modality $m$ routed to an opposite-modality ($\neg m$) expert $i$, the penalty $\lambda_i^m(t)$ is assigned a value of $\lambda_{\text{fus}}(t)$ if $i \in \mathcal{E}_{\text{fus}}^{\neg m}$, and $\lambda_{\text{spec}}(t)$ if $i \in \mathcal{E}_{\text{spec}}^{\neg m}$. As in the warmup phase, for same-modality experts $i \in \mathcal{E}^m$, $\lambda_i^m(t) = \lambda^m = 0$. We rapidly decrease the fusion penalty $\lambda_{\text{fus}}(t)$ to $\lambda^{\text{fus}}_{\min} = 0$ by training step $T^{\text{fus}}_{\text{relax}}$ to expose fusion experts to both modalities. Concurrently, the specialist penalty $\lambda_{\text{spec}}(t)$ decays more slowly until step $T^{\text{spec}}_{\text{relax}} > T^{\text{fus}}_{\text{relax}}$ to a positive residual value $\lambda^{\text{spec}}_{\min} > 0$, ensuring that specialists remain dedicated to their own modality. During the relaxation phase ($T_{\text{warm}} < t \le T^{\ast}_{\text{relax}}$), the penalty follows a cosine decay schedule for $\ast \in \{\text{fus}, \text{spec}\}$, as illustrated in Figure~\ref{fig:cosine}:
\begin{equation}
\lambda_{\ast}(t) = \lambda^\ast_{\min} + \frac{\lambda_0 - \lambda^\ast_{\min}}{2} \left[ 1 + \cos(\pi \tilde{t}) \right],
\end{equation}
where $t$ is the current training-step index, $T_{\text{warm}}$ and $T^{\ast}_{\text{relax}}$ are phase-end indices, and $\tilde{t} = (t - T_{\text{warm}})/(T^{\ast}_{\text{relax}} - T_{\text{warm}})$.

\paragraph{Penalty-Adaptive Auxiliary-Loss-Free Balancing.}
Standard auxiliary-loss-free balancing \cite{DBLP:journals/corr/abs-2412-19437} assumes a uniform target distribution of tokens across experts. However, because our routing curriculum intentionally enforces a non-uniform distribution via the penalties $\lambda_i^m(t)$, applying a standard balancing bias actively counteracts these penalties. This competition between mechanisms leads to an unbounded accumulation of the bias.

To resolve this conflict and prevent unbounded bias growth, we introduce a dynamic, penalty-adaptive target distribution. The intended routing probability for expert $i$ is formulated as a softmax over the negative penalties, and its overall target frequency $\tau_i(t)$ is computed as the expected routing probability across modalities:
\begin{equation}
\tau_i(t) = \sum_{m} \rho^m(t) \cdot \mathrm{softmax}_i(-\boldsymbol{\lambda}^m(t)),
\end{equation}
where $\boldsymbol{\lambda}^m(t) \in \mathbb{R}^E$ is the vector of penalties across all experts for modality $m$, and $\rho^m(t)$ is the exponential moving average (EMA) of the fraction of tokens belonging to modality $m$. We then update the balancing bias to correct for empirical deviations from this target:
\begin{equation}
b_i(t+1) = \gamma b_i(t) + \alpha \left( \tau_i(t) - u_i(t) \right),
\end{equation}
where $u_i(t)$ is the actual fraction of tokens routed to expert $i$, $\alpha$ controls the correction strength, and $\gamma \lesssim 1$ is a decay factor that strictly bounds the maximum bias magnitude. During inference, the accumulated balancing bias $b_i$ is frozen and retained as a static parameter.

\paragraph{Spatial Routing Signals.}
As $\lambda_i^m(t)$ decays and the probabilistic routers (governed by $\mathbf{W}_r$) assume control during differential boundary relaxation, providing explicit spatial signals can further improve routing performance \cite{DBLP:journals/jmlr/FedusZS22}, as these signals inherently serve as modality indicators. To this end, we incorporate $g(\mathbf{x})$, a spatial feature vector comprising object dimensions and relative positions to neighboring objects. For linguistic tokens, this spatial vector is simply set to zero. Furthermore, if a task does not require spatial reasoning, the router can learn to ignore $g(\mathbf{x})$ and rely entirely on the token embeddings $\mathbf{x}$. This spatial signal is particularly beneficial for embodied tasks, where object geometry and spatial layout are critical for robotic manipulation.

\begin{table*}[tp]
\centering
    \caption{\textbf{Performance comparison on 3D vision-language tasks.} The abbreviations C, B-4, M, R, Sim, and EM@1 denote CIDEr, BLEU-4, METEOR, ROUGE, Sentence Similarity, and Exact Match, respectively. Act (Total) Params denote activated (total) parameters.
    }
    \setlength{\tabcolsep}{4pt}
    \resizebox{1.0\textwidth}{!}{%
    \begin{tabular}{l | c | ccccc | ccccc | c}
    \Xhline{1.2pt}

    \multicolumn{1}{c|}{} & \multicolumn{1}{c|}{\textbf{Act (Total)}} & \multicolumn{5}{c|}{\textbf{Scan2Cap}} & \multicolumn{5}{c|}{\textbf{ScanQA}} & \multicolumn{1}{c}{\textbf{SQA3D}}\\
        & \textbf{Params}
        & C & B-4 & M & R & Sim
        & C & B-4 & M & R & EM@1
        & EM@1 \\

        \hline

        \textit{\textbf{Task-Specific Models}} &&&&&&&&&&& \\
        Scan2Cap (GPT-3) \cite{DBLP:conf/cvpr/ChenGNC21} & \NA & 35.2 & 22.4 & 21.4 & 43.5 & \NA & \NA & \NA & \NA & \NA & \NA & 41.0 \\
        3DJCG \cite{DBLP:conf/cvpr/CaiZZSX22} & \NA & 47.7 & 31.5 & 24.3 & 51.8 & \NA & \NA & \NA & \NA & \NA & \NA & \NA \\
        Vote2Cap-DETR \cite{DBLP:conf/cvpr/ChenZCLYC23} & \NA & 61.8 & 34.5 & 26.2 & 54.4 & \NA & \NA & \NA & \NA & \NA & \NA & \NA \\
        ScanRefer+MCAN \cite{DBLP:conf/eccv/ChenCN20} & \NA & \NA & \NA & \NA & \NA & \NA & 55.4 & 7.9 & 11.5 & 30.0 & 18.6 & \NA \\
        ClipBERT \cite{DBLP:conf/cvpr/LeiLZGBB021} & \NA & \NA & \NA & \NA & \NA & \NA & \NA & \NA & \NA & \NA & \NA & 43.3 \\
        ScanQA \cite{DBLP:conf/cvpr/AzumaMKK22} & \NA & \NA & \NA & \NA & \NA & \NA & 64.9 & 10.1 & 13.1 & 33.3 & 21.1 & 47.2 \\
            \hline

        \textit{\textbf{2D VLMs}} &&&&&&&&&&& \\
        LLaVA-NeXT \cite{DBLP:journals/corr/abs-2304-08485} & 7B &
            36.0  &  20.4  &  14.2  &  49.5  &  37.1  &  48.6  &  6.3  &  9.8  &  22.3  &  14.6  &  23.5
            \\
        MoE-LLaVA \cite{DBLP:journals/corr/abs-2401-15947} & 2.2B (3.1B) &
            27.4  &  18.1  &  12.1  &  40.5  &  32.2  &  42.5  &  5.0  &  7.7  &  21.8  &  11.3  &  22.3
            \\
            \hline

        \textit{\textbf{3D VLMs}} &&&&&&&&&&& \\

        3D-VisTA \cite{DBLP:conf/iccv/ZhuMCD0023} & \NA &
            66.9 & 34.0 & 27.1 & 54.3 & 53.8 & 69.6 & 10.4 & 13.9 & 35.7 & 22.4 & 48.5 \\

        3D-LLM (FlanT5) \cite{DBLP:conf/nips/HongZCZDCG23} & 3B &
            \NA & \NA & \NA & \NA & \NA & 69.4 & 12.0 & 14.5 & 35.7 & 20.5 & \NA \\

        LEO \cite{DBLP:conf/icml/HuangYMLLW0ZJ024} & 7B &
            72.4 & 38.2 & 27.9 & 58.1 & 55.3 & 101.4 & 13.2 & 20.0 & 49.2 & 24.5 & 50.0
            \\
        Chat-Scene \cite{DBLP:conf/nips/HuangC0HXWLCZPZ24} & 7B &
            77.1  &  36.3  &  28.0  &  58.1  &  \NA  &  87.7  &  14.3  &  18.0  &  41.6  &  21.6  &  54.6
            \\
        Spatial 3D-LLM \cite{DBLP:conf/icmcs/WangLXQYMLZ25} & 7B &
            72.2  &  34.6  &  23.1  &  54.3  &  \NA  &  82.5  &  13.9  &  16.8  &  39.1  &  \NA  &  46.2
            \\
        3D-LLaVA \cite{DBLP:conf/cvpr/DengHJWDR25} & 7B &
            78.8  &  36.9  &  27.1  &  57.7  &  \NA  &  92.6  &  17.1  &  18.4  &  43.1  &  \NA  &  54.5
            \\

            \hline

        \textit{\textbf{Ours}} &&&&&&&&&&& \\
        3D-MoE \texttt{1B}$\times 8 \mid \text{Top-}2$ & 1.6B (4.1B) &
            68.6  &  34.3  &  25.5  &  54.2  &  49.8  &  93.5  &  11.8  &  16.4  &  44.5  &  21.7  &  48.4
            \\

        3D-MoE \texttt{1.5B}$\times 8 \mid \text{Top-}2$ & 2.1B (5.6B) &
            71.8  &  35.7  &  26.0  &  56.6  &  50.3  &  97.2  &  14.6  &  17.5  &  46.1  &  23.9  &  52.0
            \\

        3D-MoE \texttt{3B}$\times 8 \mid \text{Top-}2$ & 4.3B (10.6B) &
            \textbf{82.2}  &  \textbf{40.6}  &  \textbf{29.4}  &  \textbf{60.7}  &  \textbf{55.5}  &  \textbf{102.2}  &  \textbf{17.8}  &  \textbf{20.2}  &  \textbf{49.6}  &  \textbf{28.2}  &  \textbf{57.8}
            \\

        \Xhline{1.2pt}
    \end{tabular}
    }
    \label{tab:3dvl}
\end{table*}

\subsection{Pose-DiT with Rectified Flow}

Spatial intelligence is a fundamental cornerstone of embodied AI. Notably, 3D vision is pivotal for robot manipulation, as it reveals critical information like 3D object geometries and locations. This makes 3D-MoE particularly well-suited for this domain. To this end, we introduce a diffusion module for predicting 6D pose actions, termed Pose-DiT. By integrating Pose-DiT as an action head, 3D-MoE is transformed into a 3D VLA for handling embodied tasks. We formulate actions as 6D poses, which describe the translation $(p_x, p_y, p_z)$ and rotation $(r_x, r_y, r_z)$ of a robot gripper \cite{DBLP:journals/corr/abs-2406-09246}. We represent each manipulation step with two poses corresponding to its pick and place targets, allowing Pose-DiT to model a structured action sequence directly in 3D space.

While flow matching \cite{DBLP:journals/corr/abs-2410-24164} is a popular choice for training VLA models, we optimize Pose-DiT using an improved variant known as rectified flow \cite{DBLP:conf/iclr/LiuG023}. The main idea is to straighten the transport path between two distributions. The benefit of straight paths is that they do not require time discretization to be simulated exactly, thereby enabling the generation of high-quality data with a single sampling step. Formally, we can view the diffusion procedure as a transport mapping problem, where the objective is to find an optimal transport map $T: \mathbb{R}^D \rightarrow \mathbb{R}^D$ such that $Z_1 = T(Z_0) \sim \mu_1$ and $Z_0 \sim \mu_0$, with observed samples from the two distributions, $X_0 \sim \mu_0$, $X_1 \sim \mu_1$. That is, $(Z_0, Z_1)$ is the transport plan, or coupling, between $\mu_0$ and $\mu_1$. In the case of 6D pose prediction, we have $D = 6$, and $\mu_0$, $\mu_1$ represent the noise and 6D pose distributions, respectively. Rectified flow is an ordinary differential equation (ODE) model with drift force $v: \mathbb{R}^D \rightarrow \mathbb{R}^D$ and time $t\in [0, 1]$:
\begin{equation}
\diff Z_t = v(Z_t, t)\diff t.
\end{equation}
To make the drift force $v$ follow the direction of the linear path from $X_0$ to $X_1$, we minimize:
\begin{equation}
\mathcal{L}_{\text{RF}} = \int_0^1 \mathbb{E} \big[ \norm{ (X_1 - X_0) -  v(X_t, t) }^2  \big] \diff t,
\label{eq:loss_rf}
\end{equation}
where $X_t = tX_1 + (1-t)X_0$ and the expectation is taken over noise samples $X_0$ and pose samples $X_1$. Owing to the non-crossing property of flows, the learned rectified flows reduce transport costs and yield straight paths. Consequently, diffusion via rectified flow requires only a single Euler discretization step, unlocking the real-time inference speeds necessary for embodied robotic control.

Furthermore, the Pose-DiT diffusion module employs an ST-DiT backbone \cite{DBLP:journals/corr/abs-2412-20404} rather than a vanilla DiT. Dedicated spatial and temporal attention blocks process pick-and-place pose sequences of shape $(L_{\text{temp}}, L_{\text{spat}}, D) = (L_{\text{temp}}, 2, 6)$ along their respective axes, and Pose-DiT is conditioned on penultimate-layer embeddings of the 3D-MoE backbone.

\begin{figure*}[t]
\centering
\captionsetup{type=table}
\centering
    \caption{\textbf{Performance comparison and ablation study on the LoHoRavens benchmark.} Models are evaluated by task success rate. Latency is reported for the diffusion process only (ms, RTX~4080~SUPER); 3D-MoE end-to-end inference, including the vision tower, takes $\sim$117\,ms. Arith. denotes Arithmetic.
    }
    \begingroup
    \setlength{\tabcolsep}{4pt}
    \resizebox{1.0\textwidth}{!}{%
    \begin{tabular}{l | *{3}{>{\Centering\arraybackslash}p{2.0cm}} | cccccc}
        \Xhline{1.2pt}

        & \multirow{2}{2.0cm}{\parbox{2.0cm}{\centering\textbf{Diffusion}\\\textbf{Type}}} &
        \multirow{2}{2.0cm}{\parbox{2.0cm}{\centering\textbf{Sampling}\\\textbf{Steps}}} &
        \multirow{2}{2.0cm}{\parbox{2.0cm}{\centering\textbf{Inference}\\\textbf{Latency}}} &
        \multicolumn{2}{c}{\textbf{Pick-\&-Place}} &
        \multirow{2}{*}{\textbf{Stack}} &
        \multicolumn{3}{c}{\textbf{Reason}} \\
        & & & & Zone & Bowl & & Spatial & Arith. & Color \\

        \hline

        \textit{\textbf{Baselines: VLM}} &  &&&&&&&& \\
        LLaVA-NeXT (2D vision) &
            \multicolumn{3}{c|}{\multirow{2}{*}{(Text-based 6D pose prediction)}} &
            0.37  &  0.40  &  0.09  &  0.46  &  0.00  &  0.14
            \\

        MoE-LLaVA (2D vision) &
            \multicolumn{3}{c|}{} &
            0.51  &  0.40  &  0.03  &  0.14  &  0.00  &  0.11
            \\

        LEO (3D vision, object point clouds) &
            \multicolumn{3}{c|}{} &
            0.69  &  0.80  &  0.23  &  0.71  &  0.17  &  0.20
            \\
            \hline

        \textit{\textbf{Baselines: VLA}} &&&&&&& \\
        OpenVLA \cite{DBLP:journals/corr/abs-2406-09246} (2D vision) &
            \multicolumn{3}{c|}{ (Direct 6D pose prediction) } &
            0.69  &  0.63  &  0.29  &  0.49  &  0.09  &  0.11
            \\

        $\pi_0$ \cite{DBLP:journals/corr/abs-2410-24164} (2D vision) &
            Flow match. & 10 & 333.55 &
            0.80  &  0.57  &  0.43  &  0.69  &  0.11  &  0.26
            \\

            \hline

        \textit{\textbf{Ours: 3D VLA}} &&&&&&& \\
        3D-MoE + Pose-DiT &
            Rec. flow & 1  &  17.08 &
            \textbf{1.00}  &  \textbf{1.00}  &  0.63  &  \textbf{1.00}  &  \textbf{0.80}  &  \textbf{0.86}
            \\

            \hline

        \textit{\textbf{Ours: Ablation Studies}} &&&&&&& \\
        W/o rectified flow &
            DDPM & 100  & 1505.21  &
            0.97  &  \textbf{1.00}  &  0.26  &  0.71  &  0.40  &  0.60
            \\

        W/o rectified flow &
            DDIM & 10  &  140.36  &
            0.37  &  0.20  &  0.09  &  0.17  &  0.00  &  0.00
            \\

        W/o rectified flow &
            Flow match. &  10  &  288.11  &
            \textbf{1.00}  &  \textbf{1.00}  &  \textbf{0.69}  &  \textbf{1.00}  &  0.71  &  0.74
            \\

        W/o diffusion &
            \multicolumn{3}{c|}{ (Direct 6D pose prediction) } &
            0.83  &  0.77  &  0.40  &  0.89  &  0.43  &  0.63
            \\

        W/o ST attention (w/ vanilla DiT) &
            Rec. flow & 1  &  2.89 &
            0.94  &  0.89  &  0.63  &  0.83  &  0.74  &  0.69
            \\

        W/o Pose-DiT (w/ MLP) &
            Rec. flow & 1  &  0.93 &
            0.40  &  0.31  &  0.00  &  0.34  &  0.00  &  0.00
            \\

        W/o spatial routing signals $g(\mathbf{x})$ &
            Rec. flow & 1  & (Default) &
            \textbf{1.00}  &  \textbf{1.00}  &  0.54  &  0.86  &  0.66  &  0.74
            \\

        W/o MoE (w/ \texttt{LLaMA3.2-1B})  &
            Rec. flow & 1  & (Default) &
            0.83  &  0.94  &  0.40  &  0.86  &  0.54  &  0.37
            \\

        W/o Stage I \& Stage II &
            Rec. flow & 1  &  (Default) &
            0.74  &  0.91  &  0.37  &  0.80  &  0.60  &  0.77
            \\

        \Xhline{1.2pt}
    \end{tabular}
    }
    \endgroup
    \label{tab:loho_ravens}

\vspace{9pt}

\captionsetup{type=figure}
    \begin{subfigure}[t]{0.48\textwidth}
        \centering
        \includegraphics[width=1.0\linewidth, trim={0.5in 9.2in 4.5in 0.5in}, clip]{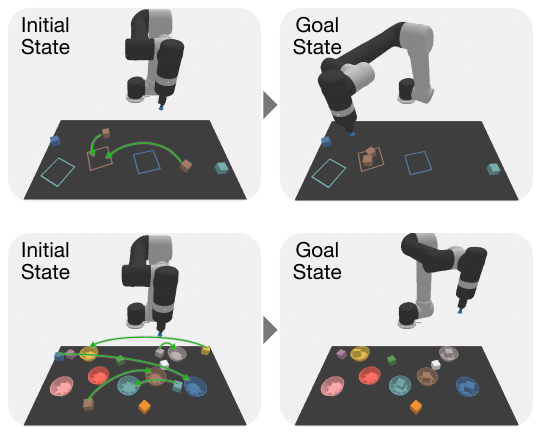}
        \caption{Arithmetic reasoning: put the blocks of an even number in the zone with the matching color.}
        \label{fig:lohoravens_ex1}
    \end{subfigure}
    \hspace{0.02\textwidth}
    \begin{subfigure}[t]{0.48\textwidth}
        \centering
        \includegraphics[width=1.0\linewidth, trim={0.5in 7.7in 4.5in 2in}, clip]{figures/example.pdf}
        \caption{Color reasoning: put the blocks in the bowls with matching colors.}
        \label{fig:lohoravens_ex2}
    \end{subfigure}
    \caption{Examples of LoHoRavens reasoning tasks successfully executed by 3D-MoE. The green arrows are included solely to highlight the actions and are not part of the input to the model.}
    \label{fig:lohoravens_example}
\end{figure*}

\section{Experiments}

\begin{table*}[t]
\centering
    \newcommand{\hypsym}[1]{\multirow{2}{*}{\makebox[0pt][l]{$#1$}\hphantom{$T^{\mathrm{spec}}_{\mathrm{relax}}$}${}\;=$}}
    \caption{\textbf{Ablation Studies.} Performance comparison of various model configurations on 3D vision-language tasks.}
    \resizebox{1.0\textwidth}{!}{%
    \begin{tabular}{ll | l | ccccc | ccccc | c}
    \Xhline{1.2pt}

    \multicolumn{2}{c|}{} & \multicolumn{1}{c|}{\multirow{2}{*}{\textbf{Description}}} & \multicolumn{5}{c|}{\textbf{Scan2Cap}} & \multicolumn{5}{c|}{\textbf{ScanQA}} & \multicolumn{1}{c}{\textbf{SQA3D}}\\
        &&
        & C & B-4 & M & R & Sim
        & C & B-4 & M & R & EM@1
        & EM@1 \\

        \hline

        \multicolumn{2}{l|}{3D-MoE \texttt{1B}$\times 8 \mid \text{Top-}2$} & Full 3D-MoE &
            68.6  &  34.3  &  25.5  &  54.2  &  49.8  &  93.5  &  11.8  &  16.4  &  44.5  &  21.7  &  48.4
            \\

        \hline

        \multicolumn{2}{l|}{\textit{\textbf{Routing Curriculum}}} &&&&&&&&&&& \\
        \multicolumn{2}{l|}{W/o differential} &
        $\lambda^{\mathrm{spec}}_{\mathrm{min}} = 0$, $T^{\text{spec}}_{\text{relax}} = T^{\text{fus}}_{\text{relax}}$ (all $\mathcal{E}_{\text{fus}}$) &
            65.4  &  30.5  &  18.6  &  53.3  &  47.4  &  91.6  &  9.7  &  14.8  &  41.1  &  18.6  &  45.8
            \\

        \multicolumn{2}{l|}{W/o relaxation} &
        $\lambda_{\text{spec}}(t) = \lambda_{\text{fus}}(t) = \lambda_0$ (all $\mathcal{E}_{\text{spec}}$) &
            67.6  &  31.9  &  23.3  &  52.2  &  47.8  &  91.3  &  10.8  &  15.2  &  43.2  &  20.0  &  44.9
            \\

        \multicolumn{2}{l|}{W/o symmetric fusion exp.} &
        Only $\mathcal{E}_{\text{fus}}^{\text{lang}}$, no $\mathcal{E}_{\text{fus}}^{\text{vis}}$ &
            68.4  &  31.0  &  23.8  &  53.7  &  49.7  &  94.5  &  12.3  &  15.3  &  43.0  &  21.9  &  48.1
            \\

        \multicolumn{2}{l|}{W/o routing curriculum} &
        Standard top-$k$ MoE &
            63.9  &  28.5  &  20.5  &  51.1  &  45.3  &  88.6  &  8.7  &  14.0  &  39.5  &  17.9  &  44.8
            \\

            \hline

        \multicolumn{2}{l|}{\textit{\textbf{Curriculum Schedule}}} &&&&&&&&&&& \\
        \hypsym{T_{\mathrm{warm}}} &
        $0$ epochs &
        \multirow{2}{*}{Default $2$ epochs} &
            64.0  &  28.8  &  21.4  &  52.5  &  46.5  &  89.2  &  8.9  &  12.9  &  40.6  &  19.8  &  41.4
            \\
        & $4$ epochs &&
            69.2  &  33.9  &  24.6  &  54.0  &  50.0  &  94.0  &  12.3  &  15.7  &  43.8  &  21.8  &  48.2
            \\
        \cline{1-2}

        \hypsym{\lambda_0} &
        $2.0$ &
        \multirow{2}{*}{Default $8.0$} &
            65.8  &  32.5  &  23.6  &  53.4  &  48.6  &  92.4  &  10.5  &  14.9  &  43.7  &  20.7  &  47.3
            \\
        & $16.0$ &&
            68.6  &  34.7  &  25.4  &  54.4  &  49.7  &  93.4  &  11.5  &  16.7  &  45.6  &  22.1  &  48.9
            \\
        \cline{1-2}

        \hypsym{T^{\mathrm{spec}}_{\mathrm{relax}}} &
        $0.5$ epoch &
        \multirow{2}{*}{Default $1$ epoch} &
            68.0  &  34.4  &  25.3  &  53.9  &  49.9  &  93.3  &  11.4  &  16.1  &  44.1  &  21.0  &  48.3
            \\
        & $2$ epochs &&
            67.5  &  35.0  &  24.9  &  54.5  &  50.1  &  93.2  &  12.2  &  15.9  &  44.5  &  21.2  &  48.6
            \\
        \cline{1-2}

        \hypsym{\lambda^{\mathrm{spec}}_{\min}} &
        $0.5$ &
        \multirow{2}{*}{Default $1.0$} &
            64.8  &  29.2  &  21.1  &  52.9  &  48.1  &  91.4  &  9.8  &  15.1  &  40.7  &  19.9  &  46.1
            \\
        & $2.0$ &&
            67.4  &  33.0  &  24.0  &  53.5  &  49.0  &  94.4  &  11.4  &  16.0  &  41.9  &  19.6  &  47.7
            \\

            \hline

        \multicolumn{2}{l|}{\textit{\textbf{Load Balancing}}} &&&&&&&&&&& \\

        \multicolumn{2}{l|}{W/o penalty-adaptive} &
        Original aux.-loss-free balancing &
            61.9  &  25.0  &  17.6  &  51.0  &  46.2  &  88.9  &  9.0  &  11.3  &  39.2  &  18.3  &  45.6
            \\

        \multicolumn{2}{l|}{W/o auxiliary-loss-free} &
        Standard load balancing loss &
            60.8  &  26.3  &  18.0  &  51.6  &  45.9  &  89.7  &  10.4  &  12.3  &  41.5  &  19.3  &  45.2
            \\

        \multicolumn{2}{l|}{W/o load balancing} &
        Load balancing disabled ($b_i=0$) &
            59.8  &  24.5  &  17.2  &  50.4  &  45.6  &  83.5  &  9.3  &  11.1  &  38.4  &  18.8  &  44.3
            \\
            \hline

        \multicolumn{2}{l|}{\textit{\textbf{Spatially Aware Routing}}} &&&&&&&&&&& \\
        \multicolumn{2}{l|}{W/o $g(\mathbf{x})$} &
        Based only on token embed. $\mathbf{x}$ &
            67.8  &  33.1  &  24.5  &  52.5  &  48.7  &  91.8  &  10.9  &  14.6  &  44.4  &  21.0  &  47.6
            \\
            \hline

        \multicolumn{2}{l|}{\textit{\textbf{Model Scaling}}} &&&&&&&&&&& \\
        \multicolumn{2}{l|}{Fewer total experts} &
        $E=4$ (total params: 2.4B) &
            56.9  &  22.9  &  18.2  &  50.6  &  44.0  &  78.8  &  8.5  &  11.9  &  36.7  &  14.5  &  43.6
            \\

        \multicolumn{2}{l|}{Fewer activated experts} &
        $k=1$ (activated params: 1.2B) &
            65.5  &  30.8  &  22.4  &  51.5  &  46.1  &  90.6  &  9.4  &  14.3  &  39.7  &  18.0  &  46.6
            \\

        \multicolumn{2}{l|}{W/o MoE} &
        Dense backbone (\texttt{LLaMA3.2-1B}) &
            31.9  &  21.9  &  15.7  &  47.4  &  41.6  &  49.6  &  6.9  &  9.2  &  26.8  &  12.1  &  26.4
            \\

        \Xhline{1.2pt}
    \end{tabular}
    }
    \label{tab:ablation}
\end{table*}

\begin{figure*}[t]
    \centering
    \includegraphics[width=\textwidth]{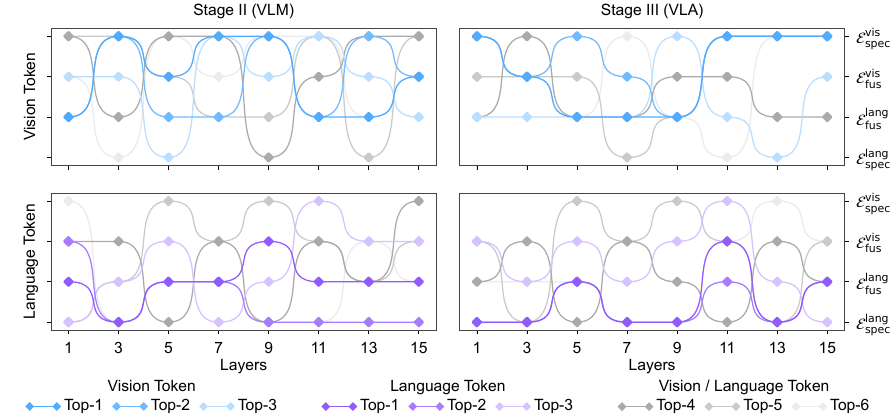}
    \caption{\textbf{Routing visualization.} Dominant routing pathways for vision tokens (top) and language tokens (bottom) in Stage~II (VLM, left) and Stage~III (VLA, right).}
    \label{fig:vis}
\end{figure*}

\subsection{Implementation Details}

We introduce a family of three 3D-MoE configurations with increasing capacities based on \texttt{LLaMA3.2-1B} \cite{llama32-meta}, \texttt{Qwen2.5-1.5B} \cite{DBLP:journals/corr/abs-2412-15115}, and \texttt{LLaMA3.2-3B}, as detailed in Table~\ref{tab:3dvl}. We use the notation ``$\times E \mid \text{Top-}k$'' to specify the MoE configuration. In the 3D vision module, we utilize PointNet++ \cite{DBLP:conf/nips/QiYSG17} to process object point clouds and a Spatial Transformer \cite{DBLP:conf/nips/ChenGTSL22} to aggregate object embeddings into scene embeddings. We optimize all models using AdamW \cite{DBLP:conf/iclr/LoshchilovH19} with a learning rate of $2 \times 10^{-5}$ and a weight decay of $0.01$. When LoRA \cite{DBLP:conf/iclr/HuSWALWWC22} is applied, we set the rank $r=64$ and the scaling factor $\alpha=128$. The models are trained for 10 epochs in Stages I and II, and 20 epochs in Stage III. For the cosine schedule of the routing curriculum, specialist warmup is applied during the sixth and seventh epochs, so $T_{\text{warm}}$ is the endpoint of the seventh epoch, with $\lambda_0 = 8.0$. We set $T^{\text{fus}}_{\text{relax}}$ to 1K steps after $T_{\text{warm}}$ and $T^{\text{spec}}_{\text{relax}}$ to one epoch after $T_{\text{warm}}$, with $\lambda^{\mathrm{spec}}_{\mathrm{min}} = 1.0$. Training is accelerated using eight NVIDIA V100 GPUs; however, thanks to the efficiency of 3D-MoE, both training and inference can be performed on a single GPU. The datasets are detailed in Appendix~\ref{sec:train_pipeline}.

\subsection{3D Vision-Language Benchmarks}

To assess spatial capabilities (e.g., 3D visual perception, spatial understanding and reasoning), we evaluate our method from Stage II on three standard 3D vision-language (3D-VL) benchmarks: Scan2Cap \cite{DBLP:conf/cvpr/ChenGNC21}, ScanQA \cite{DBLP:conf/cvpr/AzumaMKK22}, and SQA3D \cite{DBLP:conf/iclr/MaYZ0LZH23}. The quantitative results are presented in Table~\ref{tab:3dvl}.

By routing multimodal data to dedicated experts, the MoE architecture effectively mitigates the modality interference inherent to dense VLMs. As a result, 3D-MoE outperforms most baseline VLMs at a comparable total parameter count. Furthermore, scaling the 3D-MoE backbone yields substantial performance improvements while requiring far fewer active parameters during inference. Finally, we evaluate the smallest 3D-MoE configuration, which is specifically optimized to serve as a backbone for embodied agents in Stage III. Because embodied tasks rely on 6D pose actions, this model prioritizes real-time inference over heavy linguistic reasoning.

\subsection{Embodied Benchmarks}
\label{sec:embodied_tasks}

To assess the impact of spatial capabilities on embodied tasks, we evaluate our model on the LoHoRavens benchmark \cite{DBLP:journals/corr/abs-2310-12020}, with quantitative results presented in Table~\ref{tab:loho_ravens}. We report performance across six tasks: two pick-and-place tasks, one stacking task, and three significantly more challenging reasoning tasks. We conduct 35 trials per task. The baseline VLMs rely on text-based 6D pose prediction (Appendix~\ref{sec:loho_details}) to execute these tasks. Additionally, Table~\ref{tab:loho_ravens} incorporates our ablation study, investigating the effects of rectified flow, model architecture, and training stages.

Because embodied tasks prioritize visual perception and real-time inference over heavy linguistic processing, the 3D-MoE \texttt{1B}$\times 8 \mid \text{Top-}2$ configuration serves as an optimal backbone. The Pose-DiT module consists of three Transformer blocks, each featuring spatial and temporal attention layers with a hidden dimension of 256 and four attention heads. Following recent practices \cite{DBLP:journals/corr/abs-2502-02013}, Pose-DiT is conditioned on the instruction and observation embeddings extracted from the penultimate layer of the 3D-MoE backbone. Specifically, the embeddings are passed through a linear projection layer and fed into Pose-DiT's cross-attention layers. Notably, while the rectified flow framework employs 10 diffusion steps during training, it requires only a single sampling step during inference. Finally, Figure~\ref{fig:lohoravens_example} illustrates examples of 3D-MoE successfully executing the two most challenging reasoning tasks.

\subsection{Ablation Studies}
\label{sec:ablation}

Table~\ref{tab:ablation} presents a series of ablation studies analyzing the design choices behind the routing curriculum and its complementary components, including a standard top-$k$ MoE trained without the curriculum and one-dimensional sweeps of the curriculum hyperparameters. We further evaluate the overall impact of the MoE architecture and model scale. Varying $T^{\mathrm{fus}}_{\mathrm{relax}}$ around its default of 1K steps has little effect on performance, so we omit it from the table. Our findings indicate that the most critical elements are the inclusion of specialist experts and penalty-adaptive auxiliary-loss-free balancing. Additionally, as shown in Table~\ref{tab:loho_ravens}, spatial routing signals are particularly beneficial for embodied tasks due to the inherently geometric nature of the domain. Finally, in conjunction with the scaling results of the 3D-MoE variants (Table~\ref{tab:3dvl}), we demonstrate that decoupling active parameters from the total parameter count improves performance at a reduced inference cost. This highlights the promise of MoE architectures for efficiently scaling MLLMs on spatial tasks.

\begin{table}[t]
\centering
\caption{\textbf{Token Routing Distribution.} The table reports the percentages of vision and language tokens routed to each of the four expert groups across 3D-VL and embodied tasks.
}
\label{tab:expert_routing_distribution}
\resizebox{\linewidth}{!}{%
\begin{tabular}{llcc}
    \toprule
    \bf Modality & \bf Expert Group & \bf 3D-VL & \bf Embodied \\
    \midrule
    \multirow{4}{*}{\makecell[l]{\textbf{Vision}\\\textbf{Tokens}}}
    & Vision Specialist ($\mathcal{E}_{\text{spec}}^{\text{vis}}$)   & 36.2 & 43.0 \\
    & Vision Fusion ($\mathcal{E}_{\text{fus}}^{\text{vis}}$)        & 32.3 & 29.3 \\
    & Language Fusion ($\mathcal{E}_{\text{fus}}^{\text{lang}}$)      & 30.7 & 27.4 \\
    & Language Specialist ($\mathcal{E}_{\text{spec}}^{\text{lang}}$) & 0.8 & 0.3 \\
    \midrule
    \multirow{4}{*}{\makecell[l]{\textbf{Language}\\\textbf{Tokens}}}
    & Vision Specialist ($\mathcal{E}_{\text{spec}}^{\text{vis}}$)   & 2.2 & 1.9 \\
    & Vision Fusion ($\mathcal{E}_{\text{fus}}^{\text{vis}}$)        & 28.2 & 29.4 \\
    & Language Fusion ($\mathcal{E}_{\text{fus}}^{\text{lang}}$)      & 32.6 & 34.8 \\
    & Language Specialist ($\mathcal{E}_{\text{spec}}^{\text{lang}}$) & 37.1 & 33.9 \\
    \bottomrule
\end{tabular}%
}

\end{table}

\subsection{Quantitative Analysis}

To investigate the efficacy of our routing curriculum, we report the expert routing distribution in Table~\ref{tab:expert_routing_distribution} and visualize the dominant routing pathways in Figure~\ref{fig:vis} for 3D-MoE \texttt{1B}$\times 8 \mid \text{Top-}2$.

Our analysis reveals that embodied tasks demand greater visual reasoning capabilities, as evidenced by a higher proportion of vision tokens being routed to vision specialists than in standard 3D-VL tasks. This finding is further corroborated by our visualization, which demonstrates that for the VLA variant in Stage III, the top two routing pathways in the final layers are directed toward vision specialists. Conversely, in 3D-VL tasks, fusion experts play a more prominent role in handling complex spatial multimodal reasoning.

\section{Conclusion}
3D-MoE builds on advancements in 3D MLLMs and MoE architectures. Unlike prior approaches that route multimodal data indiscriminately, our routing curriculum unifies the stability of deterministic routing with the flexibility of probabilistic routing, progressively training routers to allocate experts in a modality- and spatial-context-aware manner. Combined with penalty-adaptive balancing and spatial routing signals, this improves performance on diverse 3D-VL benchmarks while activating only a subset of parameters, decoupling total capacity from per-token computation. Furthermore, by integrating the Pose-DiT action head under a rectified flow framework, we extend 3D-MoE into a 3D VLA that predicts continuous pick-and-place poses on LoHoRavens in a single sampling step, enabling real-time inference. Together, these results show that a single sparse architecture, cultivated by our routing curriculum, can serve both 3D vision-language tasks and robot manipulation, connecting 3D perception, reasoning, and embodied action.

\section*{Limitations}
Due to computational constraints, we were unable to comprehensively compare a wider variety of deterministic and probabilistic routing paradigms \cite{DBLP:journals/corr/abs-2503-07137}. For embodied tasks, we also plan to expand the scope of evaluated benchmarks \cite{DBLP:journals/corr/abs-2307-15818}. While real-world robotic experiments are highly valuable, collecting real-world training data requires immense time for manual demonstrations \cite{DBLP:journals/corr/abs-2310-08864}. Additionally, although point clouds serve as a common and robust 3D data format, future work could consider more sophisticated 3D representations, such as 3D/4D Gaussians \cite{DBLP:journals/tog/KerblKLD23, DBLP:journals/corr/abs-2603-09632, ma-etal-2026-4dgswam}.

\section*{Acknowledgments}
The research presented in this paper was partially supported by the Research Grants Council of the Hong Kong Special Administrative Region, China (CUHK 2300246, RGC C1043-24G), (CUHK 14203425, RGC GRF 2151317), and CUHK 7010870.

\clearpage

\appendix
\section{Preliminaries}

3D-MoE unifies 3D vision-language tasks and embodied manipulation tasks. In both scenarios, the model receives a multimodal input comprising a 3D scene representation $\mathbf{x}_v \in \mathbb{R}^{L_v \times d}$ (e.g., from a point cloud) and a natural language instruction $\mathbf{x}_l \in \mathbb{R}^{L_l \times d}$. Here, $L_v$ and $L_l$ denote the sequence lengths of the visual tokens and language tokens, respectively, and $d$ represents the hidden embedding dimension of the Transformer backbone. These sequences are concatenated to form the model input $\mathbf{x} = [\mathbf{x}_v; \mathbf{x}_l] \in \mathbb{R}^{(L_v + L_l) \times d}$.

Let $f(\cdot)$ denote the 3D-MoE transformer backbone. Given the input $\mathbf{x}$, the backbone produces a sequence of contextualized hidden states $\mathbf{h} = f(\mathbf{x}) \in \mathbb{R}^{(L_v + L_l) \times d}$.

For 3D-VL tasks, specific hidden states are projected by a language generation head $\mathcal{H}$ to generate a textual response $\mathbf{y}$:
\begin{equation}
\mathbf{y} = \mathcal{H}(\mathbf{h}) = \mathcal{H}(f(\mathbf{x})).
\end{equation}

For embodied tasks, the model serves as a policy $\pi(\cdot)$ to predict a sequence of 6D poses $\mathbf{A} \in \mathbb{R}^{L_{\text{temp}} \times L_{\text{spat}} \times D}$. This is achieved using the Pose-DiT diffusion head, denoted as $\phi$:
\begin{equation}
\mathbf{A} = \pi(\mathbf{x}) = \phi(\mathbf{h}) = \phi(f(\mathbf{x})).
\end{equation}
The \textit{temporal} axis $L_{\text{temp}}$ denotes the sequential steps of the task, while the \textit{spatial} axis $L_{\text{spat}}$ encodes the start and end 6D poses for each step, with $D=6$ representing 3D translation and three Euler angles.

\begin{table}[t!]
    \centering
    \caption{Notations.}
    \label{tab:notations}
    \resizebox{\columnwidth}{!}{%
    \begin{tabular}{cl}
        \toprule
        \textbf{Symbol} & \textbf{Description} \\
        \midrule

        \multicolumn{2}{c}{\textit{\textbf{3D Vision Encoder \& Language Tokenization}}} \\
        $\psi(\cdot)$ & Point cloud encoder for object-level feature extraction \\
        $\mathcal{S}(\cdot)$ & Spatial Transformer for aggregating scene-level embeddings \\
        $\mathbf{P}$ & Raw input scene consisting of a set of objects $\{\mathbf{O}_1, \dots, \mathbf{O}_{N_O}\}$ \\
        $\mathbf{O}_i$ & $i$-th object point cloud, $\mathbf{O}_i \in \mathbb{R}^{M \times 6}$ \\
        $\mathbf{h}_i$ & Object-level features for the $i$-th object, $\mathbf{h}_i \in \mathbb{R}^{d}$ \\
        $N_O$ & Number of objects in the scene \\
        $M$ & Number of points per object \\
        $\mathbf{x}_v$ & Visual token sequence, $\mathbf{x}_v \in \mathbb{R}^{L_v \times d}$ \\
        $\mathbf{x}_l$ & Language instruction token sequence, $\mathbf{x}_l \in \mathbb{R}^{L_l \times d}$ \\
        $\mathbf{x}$ & Concatenated model input, $\mathbf{x} = [\mathbf{x}_v; \mathbf{x}_l]$ \\
        $L_v$ & Sequence length for visual tokens \\
        $L_l$ & Sequence length for language tokens \\

        \midrule

        \multicolumn{2}{c}{\textit{\textbf{3D-MoE Backbone}}} \\
        $f(\cdot)$ & 3D-MoE Transformer backbone \\
        $\mathbf{h}$ & Contextualized hidden states output by the backbone, $\mathbf{h} = f(\mathbf{x})$ \\
        $d$ & Hidden embedding dimension of the Transformer \\
        $E$ & Total number of experts in the MoE layer \\
        $k$ & Number of active experts selected per token \\
        $\mathcal{K}$ & Subset of indices of the $k$ experts with the highest routing logits \\
        $\mathcal{E}_i(\mathbf{x})$ & Output embedding of the $i$-th expert FFN \\
        $\mathbf{W}_r$ & Learnable routing projection matrix, $\mathbf{W}_r \in \mathbb{R}^{(d+S) \times E}$ \\
        $t$ & Current training step \\
        $\mathbf{z}(\mathbf{x}, t)$ & Vector of routing logits at training step $t$ \\
        $z_i(\mathbf{x}, t)$ & Overall routing logit for expert $i$ \\
        $m$ & Modality of a token, $m \in \{\text{lang}, \text{vis}\}$ \\
        $S$ & Dimensionality of the spatial routing signal vector \\
        $g(\mathbf{x})$ & Spatial routing signal vector, $g(\mathbf{x}) \in \mathbb{R}^S$ \\
        $b_i(t)$ & Penalty-adaptive balancing bias at training step $t$ \\
        $\lambda_i^m(t)$ & Boundary-crossing penalty for a token of modality $m$ routed to expert $i$ \\
        $\boldsymbol{\lambda}^m(t)$ & Vector of boundary-crossing penalties across all $E$ experts \\
        $\lambda_0$ & Initial penalty value during the warmup phase \\
        $\lambda^\ast_{\min}$ & Minimum residual penalty value after boundary relaxation \\
        $\mathcal{E}_{\text{spec}}$ & Set of specialist experts (with subsets $\mathcal{E}_{\text{spec}}^{\text{lang}}, \mathcal{E}_{\text{spec}}^{\text{vis}}$) \\
        $\mathcal{E}_{\text{fus}}$ & Set of fusion experts (with subsets $\mathcal{E}_{\text{fus}}^{\text{lang}}, \mathcal{E}_{\text{fus}}^{\text{vis}}$) \\
        $T_{\text{warm}}$ & Training-step index at which specialist warmup ends \\
        $T^{\ast}_{\text{relax}}$ & Training-step index at which relaxation ends for $\ast \in \{\text{fus}, \text{spec}\}$ \\
        $\tilde{t}$ & Normalized training step progress for cosine decay schedules \\
        $\tau_i(t)$ & Target frequency (intended routing probability) for expert $i$ \\
        $u_i(t)$ & Actual fraction (utilization) of tokens routed to expert $i$ \\
        $\rho^m(t)$ & EMA fraction of tokens from modality $m$ \\
        $\alpha, \gamma$ & Correction strength and decay factor for balancing bias \\

        \midrule

        \multicolumn{2}{c}{\textit{\textbf{Language Generation Head}}} \\
        $\mathcal{H}(\cdot)$ & Language generation head \\
        $\mathbf{y}$ & Generated textual response \\
        $y_j$ & Text token at sequence step $j$ \\
        $L_y$ & Sequence length of the generated textual response \\

        \midrule

        \multicolumn{2}{c}{\textit{\textbf{6D Action Diffusion Head: Pose-DiT}}} \\
        $\phi(\cdot)$ & Pose-DiT diffusion action head \\
        $\pi(\cdot)$ & Overall policy function, $\pi(\mathbf{x}) = \phi(f(\mathbf{x}))$ \\
        $\mathbf{A}$ & Generated actions (6D poses), $\mathbf{A} \in \mathbb{R}^{L_{\text{temp}} \times L_{\text{spat}} \times D}$ \\
        $L_{\text{temp}}$ & Temporal length of action sequence (sequential task steps) \\
        $L_{\text{spat}}$ & Spatial length of action sequence (start and end poses) \\
        $D$ & Pose dimensionality, with $D=6$ (3D translation and three Euler angles) \\
        $p_x, p_y, p_z$ & Translation components of a 6D pose \\
        $r_x, r_y, r_z$ & Rotation components of a 6D pose \\
        $t$ & Continuous time for rectified flow ODE, $t \in [0, 1]$ \\
        $T$ & Optimal transport map, $T: \mathbb{R}^D \rightarrow \mathbb{R}^D$ \\
        $X_0, X_1$ & Observed samples from distributions $\mu_0$ and $\mu_1$ \\
        $Z_0, Z_1$ & State components at time $t=0, 1$ in the transport plan \\
        $Z_t$ & State component at time $t$ along the transport path \\
        $X_t$ & Interpolated state between $X_0$ and $X_1$ \\
        $v(\cdot, t)$ & Drift force (velocity field) of the ODE \\
        $\mu_0$ & Noise distribution ($t=0$) \\
        $\mu_1$ & 6D pose distribution ($t=1$) \\

        \midrule

        \multicolumn{2}{c}{\textit{\textbf{Training Objectives}}} \\
        $\mathcal{L}_{\text{CE}}$ & Cross-Entropy loss for language generation \\
        $\mathcal{L}_{\text{RF}}$ & Rectified flow objective for embodied tasks \\

        \bottomrule
    \end{tabular}
    }
\end{table}

\begin{table*}
\centering
    \caption{Model configurations.}
    \small
    \setlength{\tabcolsep}{3pt}
    \resizebox{\textwidth}{!}{%
    \begin{tabular}{@{}l|ccc|cccccc|cccc@{}}
    \Xhline{0.8pt}
         & & & \textbf{MoE}
         & & & & & \textbf{FFN} &
         & \textbf{FLOPs} & \textbf{Train-} & \textbf{Acti-} & \\
         & \textbf{Experts} & \textbf{Top-$k$} & \textbf{layers}
         & \textbf{Vocab.} & \textbf{Width} & \textbf{Layers} & \textbf{FFN} & \textbf{factor} & \textbf{Heads}
         & {\normalfont (2048 ctx)} & \textbf{able} & \textbf{vated} & \textbf{Total} \\
        \hline

        LEO &
            -- & -- & -- & 32000 & 4096 & 32 & 11008 & 3 & 32 & 14.3G & 40M & 6.7B & 6.7B \\
            \hline

        \textit{\textbf{3D-MoE}} &&&&&&&&&&&&& \\
        \texttt{1B}$\times 8 \mid \text{Top-}2$ &
            8 & 2 & 8 & 128256 & 2048 & 16 & 8192 & 3 & 32 & 3.5G & 155M & 1.6B & 4.1B \\

        \texttt{1.5B}$\times 8 \mid \text{Top-}2$ &
            8 & 2 & 14 & 151936 & 1536 & 28 & 8960 & 3 & 12 & 4.6G & 272M & 2.1B & 5.6B \\

        \texttt{3B}$\times 8 \mid \text{Top-}2$ &
            8 & 2 & 14 & 128256 & 3072 & 28 & 8192 & 3 & 24 & 9.2G & 310M & 4.3B & 10.6B \\
        \Xhline{0.8pt}
    \end{tabular}%
    }
    \label{tab:specs}
\end{table*}

\setcounter{dbltopnumber}{1}
\begin{table*}[t]
\centering
    \caption{Dataset statistics.}
    \small
    \setlength{\tabcolsep}{6pt}
    \begin{tabularx}{\textwidth}{c|>{\centering\arraybackslash}X>{\centering\arraybackslash}X>{\centering\arraybackslash}X>{\centering\arraybackslash}X>{\centering\arraybackslash}X}
    \Xhline{0.8pt}

    &  \textbf{Task}  &  \textbf{3D data}  &  \textbf{Sequences}  &   \makecell{\textbf{Tokens}\\{\normalfont (prefix+response)}}  & \makecell{\textbf{Tokens}\\{\normalfont (response)}} \\
        \hline

        \multirow[c]{3}{*}{Stage I}
            & Object captioning      & Objaverse        &  660K  &  27M  &  10M \\
            & Object referring       & ScanNet, 3RScan  &  354K  &  39M  &  15M \\
            & Scene captioning       & 3RScan           &   20K  &  4.4M  &  3.3M \\
            \hline

        \multirow[c]{4}{*}{Stage II}
            & 3D captioning          & ScanNet          &   37K  &  3M  &  821K \\
            & 3D QA                  & ScanNet, 3RScan  &   83K  &  4M  &  177K \\
            & 3D dialogue            & 3RScan           &   11K  &  8.3M  &  1.1M \\
            & Task planning          & 3RScan           &   14K  &  2.7M  &  1.9M \\
            \hline

        Stage III
            & Manipulation           & LoHoRavens       &   106K &  18M   &  4M \\

        \Xhline{0.8pt}
    \end{tabularx}
    \label{tab:datasets}
\end{table*}

\begin{figure*}[t]
    \centering
    \begin{subfigure}[t]{0.42\textwidth}
        \centering
        \includegraphics[width=1.0\linewidth, trim={0.5in 9.2in 4.5in 0.5in}, clip]{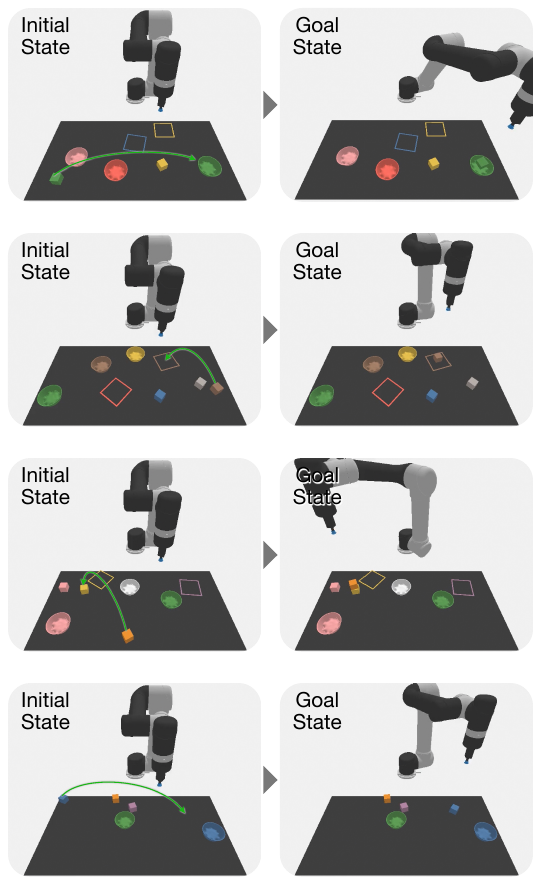}
        \caption{Pick (Bowl): pick up the green block and place it in the green bowl.}
        \label{fig:lohoravens_more_ex1}
    \end{subfigure}
    \hspace{0.08\textwidth}
    \begin{subfigure}[t]{0.42\textwidth}
        \centering
        \includegraphics[width=1.0\linewidth, trim={0.5in 7.7in 4.5in 2in}, clip]{figures/more_examples.pdf}
        \caption{Pick (Zone): pick up the brown block and place it in the brown zone.}
        \label{fig:lohoravens_more_ex2}
    \end{subfigure}
    \par\vspace{0.3cm}
    \begin{subfigure}[t]{0.42\textwidth}
        \centering
        \includegraphics[width=1.0\linewidth, trim={0.5in 6.2in 4.5in 3.5in}, clip]{figures/more_examples.pdf}
        \caption{Stacking: pick up the orange block and place it on the yellow block.}
        \label{fig:lohoravens_more_ex3}
    \end{subfigure}
    \hspace{0.08\textwidth}
    \begin{subfigure}[t]{0.42\textwidth}
        \centering
        \includegraphics[width=1.0\linewidth, trim={0.5in 4.7in 4.5in 5in}, clip]{figures/more_examples.pdf}
        \caption{Spatial reasoning: pick up the blue block in the top-left area and place it in the top-right area.}
        \label{fig:lohoravens_more_ex4}
    \end{subfigure}

    \caption{More examples of LoHoRavens tasks. The green arrows are included solely to highlight the actions and are not part of the input to the model.}
    \label{fig:lohoravens_more_example}
    \vspace{-10pt}
\end{figure*}

\begin{table*}[t]
\centering
    \caption{\textbf{End-to-end efficiency} on a single RTX~4080~SUPER (fp16). Latency is measured per action: one VLA sample or one text-based 6D pose step for VLMs. Throughput is actions per second.
    }
    \small
    \setlength{\tabcolsep}{5pt}
    \begin{tabularx}{\textwidth}{l|>{\centering\arraybackslash}X|c|>{\centering\arraybackslash}X>{\centering\arraybackslash}X>{\centering\arraybackslash}X}
        \Xhline{0.8pt}
        & \makecell[t]{\textbf{Activated}\\\textbf{(Total)}} & \makecell[t]{\textbf{Action}} & \makecell[t]{\textbf{Latency}\\\textbf{(ms)}} & \makecell[t]{\textbf{Throughput}} & \makecell[t]{\textbf{Peak}\\\textbf{Memory}} \\
        \hline

        \textit{\textbf{Baselines: VLM}} &&&&& \\
        LLaVA-NeXT &
            7B & 1 text step (94 tok) & $4183 \pm 37$ & 0.24\,Hz & 14.2\,GB \\
        MoE-LLaVA &
            2.2B (3.1B) & 1 text step (94 tok) & $4234 \pm 32$ & 0.24\,Hz & 6.7\,GB \\
        LEO &
            7B & 1 text step (80 tok) & $3545 \pm 24$ & 0.28\,Hz & 16.1\,GB \\
            \hline

        \textit{\textbf{Baselines: VLA}} &&&&& \\
        OpenVLA &
            7.5B & 1 forward pass & $114 \pm 3$ & 8.8\,Hz & 14.2\,GB \\
        $\pi_0$ &
            3.3B & Flow match., 10 & $390 \pm 19$ & 2.6\,Hz & 7.4\,GB \\
            \hline

        \textit{\textbf{Ours: 3D VLA}} &&&&& \\
        3D-MoE + Pose-DiT &
            1.6B (4.1B) & Rec.\ flow, 1 & $117 \pm 18$ & 8.5\,Hz & 7.7\,GB \\

        \Xhline{0.8pt}
    \end{tabularx}
    \label{tab:efficiency}
\end{table*}

\section{Model Details}

We show the detailed configurations of different 3D-MoE models in Table~\ref{tab:specs}. The state-of-the-art model for 3D vision-language tasks, LEO, is also included for comparison. ``FFN factor'' is the number of linear layers used in an FFN layer.

Following MoE-LLaVA \cite{DBLP:journals/corr/abs-2401-15947}, we interleave MoE blocks with the original Transformer blocks. This strategy strikes a balance between leveraging the benefits of the MoE architecture and controlling the overall model size. The size of the base LLM and the number of experts determine the number of activated and total parameters.

The inference FLOPs per token are estimated based on the autoregressive decoding phase for a single generated token. The calculation accounts for all major matrix multiplications, including query, key, value, and output projections in self-attention, and specifically models Grouped Query Attention (GQA) where applicable. For the attention mechanism, we compute the cost of attention against a cached context sequence length of $N=2048$, avoiding the quadratic cost of the prefill phase. For feed-forward networks (FFNs), 3D-MoE uses a SwiGLU structure with three linear projections. In Mixture-of-Experts (MoE) layers, the computation strictly includes the router projection and the dense FFN cost multiplied by the number of active experts ($k$), reflecting perfect sparse routing. We include the final vocabulary logits projection, but omit negligible memory-bound element-wise operations such as RMSNorm, non-linear activations, residual connections, softmax, and the Top-$k$ routing selection.

\paragraph{3D Vision Encoder.}
In the 3D vision module, we utilize PointNet++~\cite{DBLP:conf/nips/QiYSG17} to process object-centric point clouds, which are subsequently aggregated by a Spatial Transformer~\cite{DBLP:conf/nips/ChenGTSL22} to form scene-level embeddings.

Specifically, the raw input scene consists of a set of $N_O$ objects $\mathbf{P} = \{\mathbf{O}_1, \dots, \mathbf{O}_{N_O}\}$. Each object $\mathbf{O}_i \in \mathbb{R}^{M \times 6}$ consists of $M$ points, capturing both geometry and color. We first acquire object-level features $\mathbf{h}_i$ using the PointNet++ backbone $\psi$:
\begin{equation}
\mathbf{h}_i = \psi(\mathbf{O}_i) \in \mathbb{R}^{d}, \quad \text{for } i=1, \dots, N_O.
\end{equation}
To capture interactions between these objects, we employ a Spatial Transformer $\mathcal{S}$ that contextualizes the object features based on their spatial locations, resulting in the final sequence of visual tokens $\mathbf{x}_v$:
\begin{equation}
\mathbf{x}_v = \mathcal{S}(\{\mathbf{h}_1, \dots, \mathbf{h}_{N_O}\}) \in \mathbb{R}^{L_v \times d}.
\end{equation}

\paragraph{Pose-DiT.}
We detail the integration between the 3D-MoE backbone $f$ and the action head $\phi$ as follows:
\begin{itemize}
    \item \textbf{Input:} The action head $\phi$ receives the full sequence of hidden states $\mathbf{h}$ extracted from the penultimate layer of the 3D-MoE backbone, comprising embeddings for both the 3D scene and the language instruction.
    \item \textbf{Output:} The module generates 6D poses containing 3D translation and three Euler angles to control the robot end-effector: $\mathbf{A} = \phi (\mathbf{h})$.
    \item \textbf{Diffusion Framework:} We use rectified flow with 10 diffusion steps during training and a single sampling step during inference.
\end{itemize}

\paragraph{Training Objectives.}
For language generation, we minimize the standard cross-entropy loss ($\mathcal{L}_{\text{CE}}$):
\begin{equation}
\mathcal{L}_{\text{CE}} = -\frac{1}{L_y} \sum_{j=1}^{L_y} \log p(y_j \mid \mathbf{x}_v, \mathbf{x}_l, y_{<j}).
\end{equation}
For embodied tasks, we optimize the rectified flow objective defined in Equation~\eqref{eq:loss_rf}.

We compare our penalty-adaptive auxiliary-loss-free balancing with a standard loss-based balancing approach under the ``W/o auxiliary-loss-free'' configuration in Table~\ref{tab:ablation}. The standard auxiliary load balancing loss~\cite{DBLP:journals/jmlr/FedusZS22} is widely used in many previous MoE models to prevent tokens from concentrating on certain experts:
\begin{align}
\mathcal{L}_{\text{balance}} &= E \sum_{i=1}^{E} \mathcal{F}_i \cdot \mathcal{G}_i, \label{eq:loss_balance}\\
\mathcal{F}_i &= \frac{1}{kL}\sum_{j=1}^{L} \mathds{1}\{i \in \mathrm{Top}\text{-}k(\mathbf{z}(\mathbf{x}_j, t))\}, \nonumber \\
\mathcal{G}_i &= \frac{1}{L}\sum_{j=1}^{L} \text{softmax}_i(\mathbf{z}(\mathbf{x}_j, t)), \nonumber
\end{align}
where $L$ is the total sequence length. $\mathcal{F}_i$ denotes the fraction of the $kL$ expert assignments received by the $i$-th expert, counting both selected experts when $k=2$, and $\mathcal{G}_i$ is its average routing weight. This loss also enables the model to automatically assign experts to different modalities.

\paragraph{Implementation Frameworks.}
The implementation of the MoE architecture relies on DeepSpeed-MoE in the DeepSpeed library (\url{https://www.deepspeed.ai/}). To scale models, we utilize ZeRO-parallelism \cite{DBLP:conf/sc/RajbhandariRRH20}. LoRA is implemented with the PEFT library (\url{https://huggingface.co/docs/peft}).

\section{Three-stage Training Pipeline}
\label{sec:train_pipeline}

We adopt a three-stage training pipeline to progressively construct 3D-MoE. This approach seeks to maximize the transfer of learned knowledge between stages, thereby enhancing performance on downstream tasks.

\paragraph{Stage I.} In this pretraining stage, we aim to align the 3D vision encoder with the LLM via a linear projection layer and execute our proposed specialization-guiding routing curriculum. The LLM backbone is initialized with weights from a pretrained LLM. We train both the vision encoder and its projection layer while freezing the LLM, which is kept as a dense model at this point to reduce computational costs. Next, we convert the dense MLLM into the 3D-MoE architecture and pretrain it under our routing curriculum. The datasets utilized at this stage are Cap3D \cite{DBLP:conf/nips/Luo0L023} and LEO-Align \cite{DBLP:conf/icml/HuangYMLLW0ZJ024}.

\paragraph{Stage II.} In this stage, we finetune the 3D-MoE model from Stage I on various 3D vision-language (3D-VL) tasks, including Scan2Cap \cite{DBLP:conf/cvpr/ChenGNC21}, ScanQA \cite{DBLP:conf/cvpr/AzumaMKK22}, SQA3D \cite{DBLP:conf/iclr/MaYZ0LZH23}, and LEO-Instruct \cite{DBLP:conf/icml/HuangYMLLW0ZJ024}. Depending on computational constraints, the model can undergo either full parameter finetuning or parameter-efficient finetuning using LoRA \cite{DBLP:conf/iclr/HuSWALWWC22}. During this process, 3D-MoE maintains EMA updates to facilitate our penalty-adaptive auxiliary-loss-free balancing.

\paragraph{Stage III.} In the final stage, we equip 3D-MoE with an additional Pose-DiT diffusion module, enabling it to handle embodied tasks. This effectively transforms 3D-MoE into a 3D VLA model \cite{DBLP:journals/corr/abs-2307-15818}. To evaluate this capability, we utilize the recent LoHoRavens benchmark \cite{DBLP:journals/corr/abs-2310-12020}, which requires generating a series of precise 6D pose actions for robot manipulation tasks, as illustrated in Figure~\ref{fig:lohoravens_example}.

\section{Dataset Details}
In Table~\ref{tab:datasets}, we summarize the datasets used in the three stages of 3D-MoE training. Stage I and Stage II use datasets compiled by LEO \cite{DBLP:conf/icml/HuangYMLLW0ZJ024}. In Stage III, the trajectories of robot manipulation are collected using the oracle policy provided by the LoHoRavens benchmark \cite{DBLP:journals/corr/abs-2310-12020}.

As specified in the LEO paper, LEO-Align and LEO-Instruct datasets are built upon previous 3D datasets, including Objaverse \cite{DBLP:conf/cvpr/DeitkeSSWMVSEKF23}, ScanNet \cite{DBLP:conf/cvpr/DaiCSHFN17}, and 3RScan \cite{DBLP:conf/iccv/WaldANTN19}. LoHoRavens provides RGB-D data, which is also converted to point clouds. The number of tokens is an estimation since LLM models use different tokenization methods.

The evaluation metrics of 3D-VL tasks comprise CIDEr \cite{DBLP:conf/cvpr/VedantamZP15}, BLEU \cite{DBLP:conf/acl/PapineniRWZ02}, METEOR \cite{DBLP:conf/wmt/LavieA07}, ROUGE \cite{lin-2004-rouge}, and Exact Match (EM@1) \cite{DBLP:conf/iclr/MaYZ0LZH23}.

We include 2D VLMs that only process images or videos. We use an overhead view with a 45-degree tilt to capture the scene in an image. As expected, 2D images cannot provide sufficient information for 3D-VL tasks. MoE-LLaVA uses its \texttt{1.8B}\bm{$\times 4 \mid \text{Top-}2$} variant.

\section{LoHoRavens Benchmark Details}
\label{sec:loho_details}

``Pick (Zone)'' and ``Pick (Bowl)'': the agent needs to place blocks into a designated zone or bowl. ``Stack'': the embodied agent needs to stack multiple blocks. ``Reason (Spatial)'' requires reasoning about spatial descriptions such as left/right, top/bottom. This benchmark does not evaluate visual occlusion, as all objects are visible.

\begin{itemize}
    \item \textbf{Embodiment:} UR5e robot arm.
    \item \textbf{Gripper/End-Effector:} Suction gripper.
    \item \textbf{Camera Configuration:} A single Front-View RGB-D camera. While LoHoRavens provides four viewpoints (front, left, right, top-down), we utilize only the front view for all models.
    \item \textbf{Rendering Frequency:} 20~Hz.
    \item \textbf{Object Masks:} Benchmark-provided instance masks used to construct object-centric point clouds.
    \item \textbf{Action Type:} Pick-and-place motion primitives defined by pairs of 6D poses.
    \item \textbf{Action Execution:} LoHoRavens utilizes high-level motion primitives. At each time step, the robot arm moves to the target ``pick'' 6D pose to grasp the object, and subsequently moves to the ``place'' 6D pose to release it. The simulator's built-in motion planners handle low-level trajectory generation and inverse kinematics (IK).
\end{itemize}

\paragraph{Inference Efficiency.}
Table~\ref{tab:efficiency} reports end-to-end measurements, complementing the diffusion-head latencies in Table~\ref{tab:loho_ravens}. On an RTX~4080~SUPER, 3D-MoE VLA inference is $12.4$\,ms (vision) + $87.7$\,ms (backbone) + $17.1$\,ms (Pose-DiT); VL decoding runs at $17.0$ tokens/s.

\paragraph{More Examples.}
We include a few more execution examples of LoHoRavens tasks in Figure~\ref{fig:lohoravens_more_example}.

\paragraph{Text-based 6D Pose Generation.} We utilize a specifically designed format for text-based 6D pose generation for the models that do not have a dedicated action prediction head. Each pose contains 3D translation and three Euler angles.

\begin{tcolorbox}[
    width=\columnwidth,
    colback=lightgray!50,
    colframe=black!75,
    arc=3mm,
    boxrule=0.5pt,
    fontupper=\ttfamily,
    left=6pt, right=6pt, top=6pt, bottom=6pt
]
\textnormal{Format:} \\
{[step\_idx]: pick [object] from [[px,~py,~pz], [w,~rx,~ry,~rz]] and place at [[px,~py,~pz], [w,~rx,~ry,~rz]].} \\
\\
\textnormal{Examples:} \\
1: pick the green block from [[0.12, 0.34, 0.56], [1.00, 0.00, 0.00, 0.00]] and place at [[0.78, -0.91, 0.23], [1.00, 0.00, 0.00, 0.00]]. \\
\\
2: pick the blue block from [[0.12, 0.34, 0.56], [1.00, 0.00, 0.00, 0.00]] and place at [[0.78, -0.91, 0.23], [1.00, 0.00, 0.00, 0.00]]. \\
\\
3: pick the purple block from [[0.12, 0.34, 0.56], [1.00, 0.00, 0.00, 0.00]] and place at [[0.78, -0.91, 0.23], [1.00, 0.00, 0.00, 0.00]].
\end{tcolorbox}

\paragraph{Benchmark Selection.} Manipulation and navigation are both common embodied tasks. However, most navigation tasks require only discrete actions, such as \textit{turn left}, \textit{turn right}, and \textit{move forward}. Therefore, we focus on manipulation tasks that require continuous 6D pose actions, such as those found in LoHoRavens \cite{DBLP:journals/corr/abs-2310-12020}.

\section{Additional Related Work}

\subsection{Multimodal Alignment Architecture}
Reflecting trends in NLP, the ability to act on instructions has emerged as a central challenge for Vision-Language Models (VLMs), spurring research into a variety of multimodal instruction-tuning methodologies. LLaMA-Adapter \cite{DBLP:journals/corr/abs-2303-16199, DBLP:journals/corr/abs-2304-15010} implements a parameter-efficient finetuning (PEFT) strategy that equips the LLaMA model \cite{DBLP:journals/corr/abs-2302-13971} to handle visual information. A more accommodating input structure that permits interleaved images and text is introduced by Kosmos-1 \cite{DBLP:journals/corr/abs-2302-14045}, whose Magneto LLM \cite{DBLP:journals/corr/abs-2210-06423} functions as a versatile interface for connecting with perception modules \cite{DBLP:conf/icml/RadfordKHRGASAM21}. Its successor, Kosmos-2 \cite{DBLP:journals/corr/abs-2306-14824}, enhances the model with grounding and referring functionalities. To achieve instruction-following, InstructBLIP \cite{DBLP:journals/corr/abs-2305-06500} leverages an instruction-aware Q-Former derived from BLIP-2's Q-Former \cite{DBLP:conf/icml/0008LSH23}. Similar to Kosmos-2, ChatSpot \cite{DBLP:journals/corr/abs-2307-09474} demonstrates proficiency in executing precise referring instructions by employing CLIP ViT \cite{DBLP:conf/icml/RadfordKHRGASAM21} and Vicuna \cite{vicuna2023}. X-LLM \cite{DBLP:journals/corr/abs-2305-04160} utilizes X2L interfaces, which are inspired by the Q-Former from BLIP-2 \cite{DBLP:conf/icml/0008LSH23}, to translate multimodal data into a format digestible by LLMs, treating the various modalities as distinct languages. A two-stage training framework is proposed by mPLUG-Owl \cite{DBLP:journals/corr/abs-2304-14178, DBLP:journals/corr/abs-2311-04257}, which links a pretrained LLM with both a visual encoder and a visual abstractor to bestow multimodal capabilities upon the LLM. Finally, Visual ChatGPT \cite{DBLP:journals/corr/abs-2303-04671} introduces a prompt manager to govern the interplay between ChatGPT and 22 different visual foundation models, enabling ChatGPT to both interpret and produce images.

As an alternative to designing complex architectures for linking components across modalities, both LLaVA \cite{DBLP:journals/corr/abs-2304-08485} and MiniGPT-4 \cite{DBLP:journals/corr/abs-2304-10592} suggest a simpler approach: connecting vision encoders to LLMs via a single linear layer. LLaVA utilizes a two-stage training process: in the first stage, it freezes the CLIP ViT vision encoder \cite{DBLP:conf/icml/RadfordKHRGASAM21}, pretrained with contrastive learning \cite{DBLP:conf/aaai/MaSHLZK23}, and the LLM while training only the linear projection layer; in the second, it finetunes the projection layer and the LLM \cite{DBLP:journals/corr/abs-2302-13971} for visual instruction following. MiniGPT-4, however, adopts a different strategy by freezing both its vision encoder (BLIP-2's ViT + Q-Former \cite{DBLP:conf/icml/0008LSH23}) and the Vicuna LLM \cite{vicuna2023}, thereby training only the linear layer. Extending the principles of MiniGPT-4, Video-LLaMA \cite{DBLP:journals/corr/abs-2306-02858} is adapted for video processing by integrating two separate branches for video and audio, with each branch containing a specialized encoder and a Q-Former similar to that in BLIP-2 \cite{DBLP:conf/icml/0008LSH23}. PandaGPT \cite{DBLP:journals/corr/abs-2305-16355} employs ImageBind \cite{DBLP:conf/cvpr/GirdharELSAJM23} to process a wide range of data types---including vision, text, audio, depth, thermal, and IMU---which are then passed to the Vicuna model \cite{vicuna2023}, also through a linear layer. PandaGPT distinguishes itself from MiniGPT-4 by using LoRA \cite{DBLP:conf/iclr/HuSWALWWC22} to train the Vicuna model in tandem with the linear layer. Inspired by these prior works, we also employ a simple linear projection layer to align the 3D vision module with the language model.

\subsection{MoE Models}
To leverage the knowledge in pretrained LLMs, recent works propose converting existing dense LLMs into MoE architectures. Representative works include LLaMA-MoE \cite{DBLP:conf/emnlp/0002QDRTH024}, MoE-LLaVA \cite{DBLP:journals/corr/abs-2401-15947}, and Uni-MoE \cite{DBLP:journals/corr/abs-2405-11273}. These works employ trainable routers instead of fixed routers, which have proven beneficial \cite{DBLP:conf/emnlp/DikkalaGMP0023}. However, these models only focus on building 2D VLMs. We propose building an MoE model with 3D vision to achieve spatial capabilities.

\subsection{3D Vision Foundation Models}
3D Vision Foundation models have also been gaining momentum, driven by the emergence of spatial intelligence and embodied AI. Lexicon3D \cite{DBLP:conf/nips/ManZBHGW24} and OpenSU3D \cite{DBLP:journals/corr/abs-2407-14279} represent recent developments in this direction. However, since PointNet++ \cite{DBLP:conf/nips/QiYSG17} is capable of providing sufficient information for the LoHoRavens benchmark, we use it to ensure a fair comparison, avoiding giving our 3D-MoE model an advantage over baseline models in terms of 3D vision encoding.

\subsection{Spatial-Temporal Transformers}
Unlike language sequences, some data possess distinct spatial and temporal dimensions, such as video data where information is distributed across both space (the visual content within each frame) and time (the sequence of frames). The ST-DiT architecture in Open-Sora \cite{DBLP:journals/corr/abs-2412-20404} draws inspiration from ViViT \cite{DBLP:conf/iccv/Arnab0H0LS21} and Latte \cite{DBLP:journals/corr/abs-2401-03048}. ViViT is a Video Vision Transformer designed for video classification, which encodes videos by extracting spatiotemporal tokens from input videos and processing them through transformer layers to model both spatial and temporal information. Latte is a Latent Diffusion Transformer for video generation that first compresses video frames into a latent space and then applies transformer blocks to model video distribution efficiently.

\subsection{Transformer-based Diffusion Models}
Image generation diffusion models, which sit alongside variational generative models \cite{DBLP:conf/naacl/MaCLSZK24}, have also transitioned from CNN backbones \cite{DBLP:conf/miccai/RonnebergerFB15} to Transformer backbones \cite{DBLP:conf/nips/VaswaniSPUJGKP17}, exemplified by Diffusion Transformer (DiT) \cite{DBLP:conf/iccv/PeeblesX23}. Following this trend, robot policies \cite{liu2024rdt, DBLP:journals/corr/abs-2407-05996} have adopted Transformer-based backbones for diffusion-based action modules.


\begin{thebibliography}{97}
\providecommand{\natexlab}[1]{#1}

\bibitem[{Arnab et~al.(2021)Arnab, Dehghani, Heigold, Sun, Lucic, and
  Schmid}]{DBLP:conf/iccv/Arnab0H0LS21}
Anurag Arnab, Mostafa Dehghani, Georg Heigold, Chen Sun, Mario Lucic, and
  Cordelia Schmid. 2021.
\newblock Vivit: {A} video vision transformer.
\newblock In \emph{{ICCV}}, pages 6816--6826. {IEEE}.

\bibitem[{Azuma et~al.(2022)Azuma, Miyanishi, Kurita, and
  Kawanabe}]{DBLP:conf/cvpr/AzumaMKK22}
Daichi Azuma, Taiki Miyanishi, Shuhei Kurita, and Motoaki Kawanabe. 2022.
\newblock Scanqa: 3d question answering for spatial scene understanding.
\newblock In \emph{{CVPR}}, pages 19107--19117. {IEEE}.

\bibitem[{Black et~al.(2024)Black, Brown, Driess, Esmail, Equi, Finn, Fusai,
  Groom, Hausman, Ichter, Jakubczak, Jones, Ke, Levine, Li{-}Bell, Mothukuri,
  Nair, Pertsch, Shi, Tanner, Vuong, Walling, Wang, and
  Zhilinsky}]{DBLP:journals/corr/abs-2410-24164}
Kevin Black, Noah Brown, Danny Driess, Adnan Esmail, Michael Equi, Chelsea
  Finn, Niccolo Fusai, Lachy Groom, Karol Hausman, Brian Ichter, Szymon
  Jakubczak, Tim Jones, Liyiming Ke, Sergey Levine, Adrian Li{-}Bell, Mohith
  Mothukuri, Suraj Nair, Karl Pertsch, Lucy~Xiaoyang Shi, and 5 others. 2024.
\newblock {\(\pi\)}\({}_{\mbox{0}}\): {A} vision-language-action flow model for
  general robot control.
\newblock \emph{CoRR}, abs/2410.24164.

\bibitem[{Brohan et~al.(2023)Brohan, Brown, Carbajal, Chebotar, Chen,
  Choromanski, Ding, Driess, Dubey, Finn, Florence, Fu, Arenas, Gopalakrishnan,
  Han, Hausman, Herzog, Hsu, Ichter, Irpan, Joshi, Julian, Kalashnikov, Kuang,
  Leal, Lee, Lee, Levine, Lu, Michalewski, Mordatch, Pertsch, Rao, Reymann,
  Ryoo, Salazar, Sanketi, Sermanet, Singh, Singh, Soricut, Tran, Vanhoucke,
  Vuong, Wahid, Welker, Wohlhart, Wu, Xia, Xiao, Xu, Xu, Yu, and
  Zitkovich}]{DBLP:journals/corr/abs-2307-15818}
Anthony Brohan, Noah Brown, Justice Carbajal, Yevgen Chebotar, Xi~Chen,
  Krzysztof Choromanski, Tianli Ding, Danny Driess, Avinava Dubey, Chelsea
  Finn, Pete Florence, Chuyuan Fu, Montse~Gonzalez Arenas, Keerthana
  Gopalakrishnan, Kehang Han, Karol Hausman, Alexander Herzog, Jasmine Hsu,
  Brian Ichter, and 35 others. 2023.
\newblock {RT-2:} vision-language-action models transfer web knowledge to
  robotic control.
\newblock \emph{CoRR}, abs/2307.15818.

\bibitem[{Cai et~al.(2022)Cai, Zhao, Zhang, Sheng, and
  Xu}]{DBLP:conf/cvpr/CaiZZSX22}
Daigang Cai, Lichen Zhao, Jing Zhang, Lu~Sheng, and Dong Xu. 2022.
\newblock 3djcg: {A} unified framework for joint dense captioning and visual
  grounding on 3d point clouds.
\newblock In \emph{{CVPR}}, pages 16443--16452. {IEEE}.

\bibitem[{Chen et~al.(2020)Chen, Chang, and
  Nie{\ss}ner}]{DBLP:conf/eccv/ChenCN20}
Dave~Zhenyu Chen, Angel~X. Chang, and Matthias Nie{\ss}ner. 2020.
\newblock Scanrefer: 3d object localization in {RGB-D} scans using natural
  language.
\newblock In \emph{{ECCV} {(20)}}, volume 12365 of \emph{Lecture Notes in
  Computer Science}, pages 202--221. Springer.

\bibitem[{Chen et~al.(2021)Chen, Gholami, Nie{\ss}ner, and
  Chang}]{DBLP:conf/cvpr/ChenGNC21}
Dave~Zhenyu Chen, Ali Gholami, Matthias Nie{\ss}ner, and Angel~X. Chang. 2021.
\newblock Scan2cap: Context-aware dense captioning in {RGB-D} scans.
\newblock In \emph{{CVPR}}, pages 3193--3203. Computer Vision Foundation /
  {IEEE}.

\bibitem[{Chen et~al.(2023{\natexlab{a}})Chen, Han, Zhao, Zhang, Shi, Xu, and
  Xu}]{DBLP:journals/corr/abs-2305-04160}
Feilong Chen, Minglun Han, Haozhi Zhao, Qingyang Zhang, Jing Shi, Shuang Xu,
  and Bo~Xu. 2023{\natexlab{a}}.
\newblock {X-LLM:} bootstrapping advanced large language models by treating
  multi-modalities as foreign languages.
\newblock \emph{CoRR}, abs/2305.04160.

\bibitem[{Chen et~al.(2022)Chen, Guhur, Tapaswi, Schmid, and
  Laptev}]{DBLP:conf/nips/ChenGTSL22}
Shizhe Chen, Pierre{-}Louis Guhur, Makarand Tapaswi, Cordelia Schmid, and Ivan
  Laptev. 2022.
\newblock Language conditioned spatial relation reasoning for 3d object
  grounding.
\newblock In \emph{NeurIPS}.

\bibitem[{Chen et~al.(2023{\natexlab{b}})Chen, Zhu, Chen, Lei, Yu, and
  Chen}]{DBLP:conf/cvpr/ChenZCLYC23}
Sijin Chen, Hongyuan Zhu, Xin Chen, Yinjie Lei, Gang Yu, and Tao Chen.
  2023{\natexlab{b}}.
\newblock End-to-end 3d dense captioning with vote2cap-detr.
\newblock In \emph{{CVPR}}, pages 11124--11133. {IEEE}.

\bibitem[{Chi et~al.(2023)Chi, Feng, Du, Xu, Cousineau, Burchfiel, and
  Song}]{DBLP:conf/rss/ChiFDXCBS23}
Cheng Chi, Siyuan Feng, Yilun Du, Zhenjia Xu, Eric Cousineau, Benjamin
  Burchfiel, and Shuran Song. 2023.
\newblock Diffusion policy: Visuomotor policy learning via action diffusion.
\newblock In \emph{Robotics: Science and Systems}.

\bibitem[{Chiang et~al.(2023)Chiang, Li, Lin, Sheng, Wu, Zhang, Zheng, Zhuang,
  Zhuang, Gonzalez, Stoica, and Xing}]{vicuna2023}
Wei-Lin Chiang, Zhuohan Li, Zi~Lin, Ying Sheng, Zhanghao Wu, Hao Zhang, Lianmin
  Zheng, Siyuan Zhuang, Yonghao Zhuang, Joseph~E. Gonzalez, Ion Stoica, and
  Eric~P. Xing. 2023.
\newblock \href {https://lmsys.org/blog/2023-03-30-vicuna/} {Vicuna: An
  open-source chatbot impressing gpt-4 with 90\%* chatgpt quality}.

\bibitem[{Collaboration et~al.(2023)Collaboration, Padalkar, Pooley, Jain,
  Bewley, Herzog, Irpan, Khazatsky, Raj, Singh, Brohan, Raffin, Wahid,
  Burgess{-}Limerick, Kim, Sch{\"{o}}lkopf, Ichter, Lu, Xu, Finn, Xu, Chi,
  Huang, Chan, Pan, Fu, Devin, Driess, Pathak, Shah, B{\"{u}}chler,
  Kalashnikov, Sadigh, Johns, Ceola, Xia, Stulp, Zhou, Sukhatme, Salhotra, Yan,
  Schiavi, Kahn, Su, Fang, Shi, Amor, Christensen, Furuta, Walke, Fang,
  Mordatch, Radosavovic, and et~al.}]{DBLP:journals/corr/abs-2310-08864}
Open~X.{-}Embodiment Collaboration, Abhishek Padalkar, Acorn Pooley, Ajinkya
  Jain, Alex Bewley, Alexander Herzog, Alex Irpan, Alexander Khazatsky, Anant
  Raj, Anikait Singh, Anthony Brohan, Antonin Raffin, Ayzaan Wahid, Ben
  Burgess{-}Limerick, Beomjoon Kim, Bernhard Sch{\"{o}}lkopf, Brian Ichter,
  Cewu Lu, Charles Xu, and 35 others. 2023.
\newblock Open x-embodiment: Robotic learning datasets and {RT-X} models.
\newblock \emph{CoRR}, abs/2310.08864.

\bibitem[{Dai et~al.(2017)Dai, Chang, Savva, Halber, Funkhouser, and
  Nie{\ss}ner}]{DBLP:conf/cvpr/DaiCSHFN17}
Angela Dai, Angel~X. Chang, Manolis Savva, Maciej Halber, Thomas~A. Funkhouser,
  and Matthias Nie{\ss}ner. 2017.
\newblock Scannet: Richly-annotated 3d reconstructions of indoor scenes.
\newblock In \emph{{CVPR}}, pages 2432--2443. {IEEE} Computer Society.

\bibitem[{Dai et~al.(2023)Dai, Li, Li, Tiong, Zhao, Wang, Li, Fung, and
  Hoi}]{DBLP:journals/corr/abs-2305-06500}
Wenliang Dai, Junnan Li, Dongxu Li, Anthony Meng~Huat Tiong, Junqi Zhao,
  Weisheng Wang, Boyang Li, Pascale Fung, and Steven C.~H. Hoi. 2023.
\newblock Instructblip: Towards general-purpose vision-language models with
  instruction tuning.
\newblock \emph{CoRR}, abs/2305.06500.

\bibitem[{DeepSeek{-}AI et~al.(2024)DeepSeek{-}AI, Liu, Feng, Xue, Wang, Wu,
  Lu, Zhao, Deng, Zhang, Ruan, Dai, Guo, Yang, Chen, Ji, Li, Lin, Dai, Luo,
  Hao, Chen, Li, Zhang, Bao, Xu, Wang, Zhang, Ding, Xin, Gao, Li, Qu, Cai,
  Liang, Guo, Ni, Li, Wang, Chen, Chen, Yuan, Qiu, Li, Song, Dong, Hu, Gao,
  Guan, Huang, Yu, Wang, Zhang, Xu, Xia, Zhao, Wang, Zhang, Li, Wang, Zhang,
  Zhang, Tang, Li, Tian, Huang, Wang, Zhang, Wang, Zhu, Chen, Du, Chen, Jin,
  Ge, Zhang, Pan, Wang, Xu, Zhang, Chen, Li, Lu, Zhou, Chen, Wu, Ye, Ye, Ma,
  Wang, Zhou, Yu, Zhou, Pan, Wang, Yun, Pei, Sun, Xiao, and
  Zeng}]{DBLP:journals/corr/abs-2412-19437}
DeepSeek{-}AI, Aixin Liu, Bei Feng, Bing Xue, Bingxuan Wang, Bochao Wu, Chengda
  Lu, Chenggang Zhao, Chengqi Deng, Chenyu Zhang, Chong Ruan, Damai Dai, Daya
  Guo, Dejian Yang, Deli Chen, Dongjie Ji, Erhang Li, Fangyun Lin, Fucong Dai,
  and 81 others. 2024.
\newblock Deepseek-v3 technical report.
\newblock \emph{CoRR}, abs/2412.19437.

\bibitem[{Deitke et~al.(2023)Deitke, Schwenk, Salvador, Weihs, Michel,
  VanderBilt, Schmidt, Ehsani, Kembhavi, and
  Farhadi}]{DBLP:conf/cvpr/DeitkeSSWMVSEKF23}
Matt Deitke, Dustin Schwenk, Jordi Salvador, Luca Weihs, Oscar Michel, Eli
  VanderBilt, Ludwig Schmidt, Kiana Ehsani, Aniruddha Kembhavi, and Ali
  Farhadi. 2023.
\newblock Objaverse: {A} universe of annotated 3d objects.
\newblock In \emph{{CVPR}}, pages 13142--13153. {IEEE}.

\bibitem[{Deng et~al.(2025)Deng, He, Jiang, Wang, Dayoub, and
  Reid}]{DBLP:conf/cvpr/DengHJWDR25}
Jiajun Deng, Tianyu He, Li~Jiang, Tianyu Wang, Feras Dayoub, and Ian~D. Reid.
  2025.
\newblock 3d-llava: Towards generalist 3d lmms with omni superpoint
  transformer.
\newblock In \emph{{CVPR}}, pages 3772--3782. Computer Vision Foundation /
  {IEEE}.

\bibitem[{Dikkala et~al.(2023)Dikkala, Ghosh, Meka, Panigrahy, Vyas, and
  Wang}]{DBLP:conf/emnlp/DikkalaGMP0023}
Nishanth Dikkala, Nikhil Ghosh, Raghu Meka, Rina Panigrahy, Nikhil Vyas, and
  Xin Wang. 2023.
\newblock On the benefits of learning to route in mixture-of-experts models.
\newblock In \emph{{EMNLP}}, pages 9376--9396. Association for Computational
  Linguistics.

\bibitem[{Fedus et~al.(2022)Fedus, Zoph, and
  Shazeer}]{DBLP:journals/jmlr/FedusZS22}
William Fedus, Barret Zoph, and Noam Shazeer. 2022.
\newblock Switch transformers: Scaling to trillion parameter models with simple
  and efficient sparsity.
\newblock \emph{J. Mach. Learn. Res.}, 23:120:1--120:39.

\bibitem[{Gao et~al.(2023)Gao, Han, Zhang, Lin, Geng, Zhou, Zhang, Lu, He, Yue,
  Li, and Qiao}]{DBLP:journals/corr/abs-2304-15010}
Peng Gao, Jiaming Han, Renrui Zhang, Ziyi Lin, Shijie Geng, Aojun Zhou, Wei
  Zhang, Pan Lu, Conghui He, Xiangyu Yue, Hongsheng Li, and Yu~Qiao. 2023.
\newblock Llama-adapter {V2:} parameter-efficient visual instruction model.
\newblock \emph{CoRR}, abs/2304.15010.

\bibitem[{Girdhar et~al.(2023)Girdhar, El{-}Nouby, Liu, Singh, Alwala, Joulin,
  and Misra}]{DBLP:conf/cvpr/GirdharELSAJM23}
Rohit Girdhar, Alaaeldin El{-}Nouby, Zhuang Liu, Mannat Singh, Kalyan~Vasudev
  Alwala, Armand Joulin, and Ishan Misra. 2023.
\newblock Imagebind one embedding space to bind them all.
\newblock In \emph{{CVPR}}, pages 15180--15190. {IEEE}.

\bibitem[{Ho et~al.(2020)Ho, Jain, and Abbeel}]{DBLP:conf/nips/HoJA20}
Jonathan Ho, Ajay Jain, and Pieter Abbeel. 2020.
\newblock Denoising diffusion probabilistic models.
\newblock In \emph{NeurIPS}.

\bibitem[{Hong et~al.(2023)Hong, Zhen, Chen, Zheng, Du, Chen, and
  Gan}]{DBLP:conf/nips/HongZCZDCG23}
Yining Hong, Haoyu Zhen, Peihao Chen, Shuhong Zheng, Yilun Du, Zhenfang Chen,
  and Chuang Gan. 2023.
\newblock 3d-llm: Injecting the 3d world into large language models.
\newblock In \emph{NeurIPS}.

\bibitem[{Hu et~al.(2022)Hu, Shen, Wallis, Allen{-}Zhu, Li, Wang, Wang, and
  Chen}]{DBLP:conf/iclr/HuSWALWWC22}
Edward~J. Hu, Yelong Shen, Phillip Wallis, Zeyuan Allen{-}Zhu, Yuanzhi Li,
  Shean Wang, Lu~Wang, and Weizhu Chen. 2022.
\newblock Lora: Low-rank adaptation of large language models.
\newblock In \emph{{ICLR}}. OpenReview.net.

\bibitem[{Huang et~al.(2024{\natexlab{a}})Huang, Chen, Wang, Huang, Xu, Wang,
  Liu, Cheng, Zhao, Pang, and Zhao}]{DBLP:conf/nips/HuangC0HXWLCZPZ24}
Haifeng Huang, Yilun Chen, Zehan Wang, Rongjie Huang, Runsen Xu, Tai Wang,
  Luping Liu, Xize Cheng, Yang Zhao, Jiangmiao Pang, and Zhou Zhao.
  2024{\natexlab{a}}.
\newblock Chat-scene: Bridging 3d scene and large language models with object
  identifiers.
\newblock In \emph{NeurIPS}.

\bibitem[{Huang et~al.(2024{\natexlab{b}})Huang, Yong, Ma, Linghu, Li, Wang,
  Li, Zhu, Jia, and Huang}]{DBLP:conf/icml/HuangYMLLW0ZJ024}
Jiangyong Huang, Silong Yong, Xiaojian Ma, Xiongkun Linghu, Puhao Li, Yan Wang,
  Qing Li, Song{-}Chun Zhu, Baoxiong Jia, and Siyuan Huang. 2024{\natexlab{b}}.
\newblock An embodied generalist agent in 3d world.
\newblock In \emph{{ICML}}. OpenReview.net.

\bibitem[{Huang et~al.(2023)Huang, Dong, Wang, Hao, Singhal, Ma, Lv, Cui,
  Mohammed, Patra, Liu, Aggarwal, Chi, Bjorck, Chaudhary, Som, Song, and
  Wei}]{DBLP:journals/corr/abs-2302-14045}
Shaohan Huang, Li~Dong, Wenhui Wang, Yaru Hao, Saksham Singhal, Shuming Ma,
  Tengchao Lv, Lei Cui, Owais~Khan Mohammed, Barun Patra, Qiang Liu, Kriti
  Aggarwal, Zewen Chi, Johan Bjorck, Vishrav Chaudhary, Subhojit Som, Xia Song,
  and Furu Wei. 2023.
\newblock Language is not all you need: Aligning perception with language
  models.
\newblock \emph{CoRR}, abs/2302.14045.

\bibitem[{Jiang et~al.(2024)Jiang, Sablayrolles, Roux, Mensch, Savary, Bamford,
  Chaplot, de~Las~Casas, Hanna, Bressand, Lengyel, Bour, Lample, Lavaud,
  Saulnier, Lachaux, Stock, Subramanian, Yang, Antoniak, Scao, Gervet, Lavril,
  Wang, Lacroix, and Sayed}]{DBLP:journals/corr/abs-2401-04088}
Albert~Q. Jiang, Alexandre Sablayrolles, Antoine Roux, Arthur Mensch, Blanche
  Savary, Chris Bamford, Devendra~Singh Chaplot, Diego de~Las~Casas, Emma~Bou
  Hanna, Florian Bressand, Gianna Lengyel, Guillaume Bour, Guillaume Lample,
  L{\'{e}}lio~Renard Lavaud, Lucile Saulnier, Marie{-}Anne Lachaux, Pierre
  Stock, Sandeep Subramanian, Sophia Yang, and 7 others. 2024.
\newblock Mixtral of experts.
\newblock \emph{CoRR}, abs/2401.04088.

\bibitem[{Ke et~al.(2024)Ke, Gkanatsios, and
  Fragkiadaki}]{DBLP:journals/corr/abs-2402-10885}
Tsung{-}Wei Ke, Nikolaos Gkanatsios, and Katerina Fragkiadaki. 2024.
\newblock 3d diffuser actor: Policy diffusion with 3d scene representations.
\newblock \emph{CoRR}, abs/2402.10885.

\bibitem[{Kerbl et~al.(2023)Kerbl, Kopanas, Leimk{\"{u}}hler, and
  Drettakis}]{DBLP:journals/tog/KerblKLD23}
Bernhard Kerbl, Georgios Kopanas, Thomas Leimk{\"{u}}hler, and George
  Drettakis. 2023.
\newblock 3d gaussian splatting for real-time radiance field rendering.
\newblock \emph{{ACM} Trans. Graph.}, 42(4):139:1--139:14.

\bibitem[{Kim et~al.(2024)Kim, Pertsch, Karamcheti, Xiao, Balakrishna, Nair,
  Rafailov, Foster, Lam, Sanketi, Vuong, Kollar, Burchfiel, Tedrake, Sadigh,
  Levine, Liang, and Finn}]{DBLP:journals/corr/abs-2406-09246}
Moo~Jin Kim, Karl Pertsch, Siddharth Karamcheti, Ted Xiao, Ashwin Balakrishna,
  Suraj Nair, Rafael Rafailov, Ethan~Paul Foster, Grace Lam, Pannag Sanketi,
  Quan Vuong, Thomas Kollar, Benjamin Burchfiel, Russ Tedrake, Dorsa Sadigh,
  Sergey Levine, Percy Liang, and Chelsea Finn. 2024.
\newblock Openvla: An open-source vision-language-action model.
\newblock \emph{CoRR}, abs/2406.09246.

\bibitem[{Lavie and Agarwal(2007)}]{DBLP:conf/wmt/LavieA07}
Alon Lavie and Abhaya Agarwal. 2007.
\newblock {METEOR:} an automatic metric for {MT} evaluation with high levels of
  correlation with human judgments.
\newblock In \emph{WMT@ACL}, pages 228--231. Association for Computational
  Linguistics.

\bibitem[{Lei et~al.(2021)Lei, Li, Zhou, Gan, Berg, Bansal, and
  Liu}]{DBLP:conf/cvpr/LeiLZGBB021}
Jie Lei, Linjie Li, Luowei Zhou, Zhe Gan, Tamara~L. Berg, Mohit Bansal, and
  Jingjing Liu. 2021.
\newblock Less is more: Clipbert for video-and-language learning via sparse
  sampling.
\newblock In \emph{{CVPR}}, pages 7331--7341. Computer Vision Foundation /
  {IEEE}.

\bibitem[{Li et~al.(2024{\natexlab{a}})Li, Johnson, Lassner, and
  Mildenhall}]{world-labs-about}
Fei-Fei Li, Justin Johnson, Christoph Lassner, and Ben Mildenhall.
  2024{\natexlab{a}}.
\newblock \href {https://www.worldlabs.ai/about} {What is spatial
  intelligence?}

\bibitem[{Li et~al.(2023)Li, Li, Savarese, and Hoi}]{DBLP:conf/icml/0008LSH23}
Junnan Li, Dongxu Li, Silvio Savarese, and Steven C.~H. Hoi. 2023.
\newblock {BLIP-2:} bootstrapping language-image pre-training with frozen image
  encoders and large language models.
\newblock In \emph{{ICML}}, volume 202 of \emph{Proceedings of Machine Learning
  Research}, pages 19730--19742. {PMLR}.

\bibitem[{Li et~al.(2025)Li, Chen, Jiang, Shi, Liu, Zhang, Deng, Xu, Ma, Zhang,
  Hu, and Zhang}]{DBLP:journals/corr/abs-2511-12609}
Yunxin Li, Xinyu Chen, Shenyuan Jiang, Haoyuan Shi, Zhenyu Liu, Xuanyu Zhang,
  Nanhao Deng, Zhenran Xu, Yicheng Ma, Meishan Zhang, Baotian Hu, and Min
  Zhang. 2025.
\newblock Uni-moe-2.0-omni: Scaling language-centric omnimodal large model with
  advanced moe, training and data.
\newblock \emph{CoRR}, abs/2511.12609.

\bibitem[{Li et~al.(2024{\natexlab{b}})Li, Jiang, Hu, Wang, Zhong, Luo, Ma, and
  Zhang}]{DBLP:journals/corr/abs-2405-11273}
Yunxin Li, Shenyuan Jiang, Baotian Hu, Longyue Wang, Wanqi Zhong, Wenhan Luo,
  Lin Ma, and Min Zhang. 2024{\natexlab{b}}.
\newblock Uni-moe: Scaling unified multimodal llms with mixture of experts.
\newblock \emph{CoRR}, abs/2405.11273.

\bibitem[{Lin et~al.(2024)Lin, Tang, Ye, Cui, Zhu, Jin, Zhang, Ning, and
  Yuan}]{DBLP:journals/corr/abs-2401-15947}
Bin Lin, Zhenyu Tang, Yang Ye, Jiaxi Cui, Bin Zhu, Peng Jin, Junwu Zhang, Munan
  Ning, and Li~Yuan. 2024.
\newblock Moe-llava: Mixture of experts for large vision-language models.
\newblock \emph{CoRR}, abs/2401.15947.

\bibitem[{Lin(2004)}]{lin-2004-rouge}
Chin-Yew Lin. 2004.
\newblock {ROUGE}: A package for automatic evaluation of summaries.
\newblock In \emph{Text Summarization Branches Out}, pages 74--81, Barcelona,
  Spain. Association for Computational Linguistics.

\bibitem[{Liu et~al.(2023{\natexlab{a}})Liu, Li, Wu, and
  Lee}]{DBLP:journals/corr/abs-2304-08485}
Haotian Liu, Chunyuan Li, Qingyang Wu, and Yong~Jae Lee. 2023{\natexlab{a}}.
\newblock Visual instruction tuning.
\newblock \emph{CoRR}, abs/2304.08485.

\bibitem[{Liu et~al.(2024)Liu, Wu, Li, Tan, Chen, Wang, Xu, Su, and
  Zhu}]{liu2024rdt}
Songming Liu, Lingxuan Wu, Bangguo Li, Hengkai Tan, Huayu Chen, Zhengyi Wang,
  Ke~Xu, Hang Su, and Jun Zhu. 2024.
\newblock Rdt-1b: a diffusion foundation model for bimanual manipulation.
\newblock \emph{arXiv preprint arXiv:2410.07864}.

\bibitem[{Liu et~al.(2023{\natexlab{b}})Liu, Gong, and
  Liu}]{DBLP:conf/iclr/LiuG023}
Xingchao Liu, Chengyue Gong, and Qiang Liu. 2023{\natexlab{b}}.
\newblock Flow straight and fast: Learning to generate and transfer data with
  rectified flow.
\newblock In \emph{{ICLR}}. OpenReview.net.

\bibitem[{Loshchilov and Hutter(2019)}]{DBLP:conf/iclr/LoshchilovH19}
Ilya Loshchilov and Frank Hutter. 2019.
\newblock Decoupled weight decay regularization.
\newblock In \emph{{ICLR} (Poster)}. OpenReview.net.

\bibitem[{Luo et~al.(2025{\natexlab{a}})Luo, Dou, Li, Wang, Yang, Tian, Li,
  Wang, Wang, Zhu, Qiao, and Dai}]{DBLP:journals/corr/abs-2507-12566}
Gen Luo, Wenhan Dou, Wenhao Li, Zhaokai Wang, Xue Yang, Changyao Tian, Hao Li,
  Weiyun Wang, Wenhai Wang, Xizhou Zhu, Yu~Qiao, and Jifeng Dai.
  2025{\natexlab{a}}.
\newblock Mono-internvl-1.5: Towards cheaper and faster monolithic multimodal
  large language models.
\newblock \emph{CoRR}, abs/2507.12566.

\bibitem[{Luo et~al.(2025{\natexlab{b}})Luo, Yang, Dou, Wang, Liu, Dai, Qiao,
  and Zhu}]{DBLP:conf/cvpr/LuoYDWLD0Z25}
Gen Luo, Xue Yang, Wenhan Dou, Zhaokai Wang, Jiawen Liu, Jifeng Dai, Yu~Qiao,
  and Xizhou Zhu. 2025{\natexlab{b}}.
\newblock Mono-internvl: Pushing the boundaries of monolithic multimodal large
  language models with endogenous visual pre-training.
\newblock In \emph{{CVPR}}, pages 24960--24971. Computer Vision Foundation /
  {IEEE}.

\bibitem[{Luo et~al.(2023)Luo, Rockwell, Lee, and
  Johnson}]{DBLP:conf/nips/Luo0L023}
Tiange Luo, Chris Rockwell, Honglak Lee, and Justin Johnson. 2023.
\newblock Scalable 3d captioning with pretrained models.
\newblock In \emph{NeurIPS}.

\bibitem[{Ma et~al.(2023{\natexlab{a}})Ma, Yong, Zheng, Li, Liang, Zhu, and
  Huang}]{DBLP:conf/iclr/MaYZ0LZH23}
Xiaojian Ma, Silong Yong, Zilong Zheng, Qing Li, Yitao Liang, Song{-}Chun Zhu,
  and Siyuan Huang. 2023{\natexlab{a}}.
\newblock {SQA3D:} situated question answering in 3d scenes.
\newblock In \emph{{ICLR}}. OpenReview.net.

\bibitem[{Ma et~al.(2024{\natexlab{a}})Ma, Wang, Jia, Chen, Liu, Li, Chen, and
  Qiao}]{DBLP:journals/corr/abs-2401-03048}
Xin Ma, Yaohui Wang, Gengyun Jia, Xinyuan Chen, Ziwei Liu, Yuan{-}Fang Li,
  Cunjian Chen, and Yu~Qiao. 2024{\natexlab{a}}.
\newblock Latte: Latent diffusion transformer for video generation.
\newblock \emph{CoRR}, abs/2401.03048.

\bibitem[{Ma et~al.(2024{\natexlab{b}})Ma, Chi, Li, Song, Zhuang, and
  King}]{DBLP:conf/naacl/MaCLSZK24}
Yueen Ma, Dafeng Chi, Jingjing Li, Kai Song, Yuzheng Zhuang, and Irwin King.
  2024{\natexlab{b}}.
\newblock {VOLTA:} improving generative diversity by variational mutual
  information maximizing autoencoder.
\newblock In \emph{{NAACL-HLT} (Findings)}, volume {NAACL} 2024 of
  \emph{Findings of {ACL}}, pages 364--378. Association for Computational
  Linguistics.

\bibitem[{Ma et~al.(2025)Ma, Chi, Wu, Liu, Zhuang, and
  King}]{ma-etal-2025-astra}
Yueen Ma, DaFeng Chi, Shiguang Wu, Yuecheng Liu, Yuzheng Zhuang, and Irwin
  King. 2025.
\newblock Astra: Efficient transformer architecture and contrastive dynamics
  learning for embodied instruction following.
\newblock In \emph{EMNLP}.

\bibitem[{Ma et~al.(2023{\natexlab{b}})Ma, Song, Hu, Li, Zhang, and
  King}]{DBLP:conf/aaai/MaSHLZK23}
Yueen Ma, Zixing Song, Xuming Hu, Jingjing Li, Yifei Zhang, and Irwin King.
  2023{\natexlab{b}}.
\newblock Graph component contrastive learning for concept relatedness
  estimation.
\newblock In \emph{{AAAI}}, pages 13362--13370. {AAAI} Press.

\bibitem[{Ma et~al.(2026{\natexlab{a}})Ma, Song, Zhuang, Hao, and
  King}]{DBLP:journals/tnn/MaSZHK26}
Yueen Ma, Zixing Song, Yuzheng Zhuang, Jianye Hao, and Irwin King.
  2026{\natexlab{a}}.
\newblock A survey on vision-language-action models for embodied {AI}.
\newblock \emph{{IEEE} Trans. Neural Networks Learn. Syst.}, 37(7):3031--3051.

\bibitem[{Ma et~al.(2026{\natexlab{b}})Ma, Xu, and King}]{ma-etal-2026-4dgswam}
Yueen Ma, Zenglin Xu, and Irwin King. 2026{\natexlab{b}}.
\newblock {4DGS-WAM:} bridging past and future with an object-centric world
  action model based on {4D} gaussian splatting.
\newblock \emph{CoRR}, abs/2608.25956.

\bibitem[{Ma et~al.(2026{\natexlab{c}})Ma, Xu, and
  King}]{DBLP:journals/corr/abs-2603-09632}
Yueen Ma, Zenglin Xu, and Irwin King. 2026{\natexlab{c}}.
\newblock {X-GS:} an extensible open framework for perceiving and thinking via
  3d gaussian splatting.
\newblock \emph{CoRR}, abs/2603.09632.

\bibitem[{Man et~al.(2024)Man, Zheng, Bao, Hebert, Gui, and
  Wang}]{DBLP:conf/nips/ManZBHGW24}
Yunze Man, Shuhong Zheng, Zhipeng Bao, Martial Hebert, Liangyan Gui, and
  Yu{-}Xiong Wang. 2024.
\newblock \href
  {http://papers.nips.cc/paper\_files/paper/2024/hash/8c67fc501a50977947c5bebbc39ca8f6-Abstract-Conference.html}
  {Lexicon3d: Probing visual foundation models for complex 3d scene
  understanding}.
\newblock In \emph{Advances in Neural Information Processing Systems 38: Annual
  Conference on Neural Information Processing Systems 2024, NeurIPS 2024,
  Vancouver, BC, Canada, December 10 - 15, 2024}.

\bibitem[{Meta(2024)}]{llama32-meta}
Meta. 2024.
\newblock \href {https://www.llama.com/models/llama-3/} {Llama 3.2:
  Revolutionizing edge ai and vision with open, customizable models}.

\bibitem[{Meta(2025)}]{llama4-meta}
Meta. 2025.
\newblock \href {https://www.llama.com/models/llama-4/} {Llama 4}.

\bibitem[{Mohiuddin et~al.(2024)Mohiuddin, Prakhya, Collins, Liu, and
  Borrmann}]{DBLP:journals/corr/abs-2407-14279}
Rafay Mohiuddin, Sai~Manoj Prakhya, Fiona~C. Collins, Ziyuan Liu, and
  Andr{\'{e}} Borrmann. 2024.
\newblock \href {https://doi.org/10.48550/ARXIV.2407.14279} {Opensu3d: Open
  world 3d scene understanding using foundation models}.
\newblock \emph{CoRR}, abs/2407.14279.

\bibitem[{Mu and Lin(2025)}]{DBLP:journals/corr/abs-2503-07137}
Siyuan Mu and Sen Lin. 2025.
\newblock A comprehensive survey of mixture-of-experts: Algorithms, theory, and
  applications.
\newblock \emph{CoRR}, abs/2503.07137.

\bibitem[{Mustafa et~al.(2022)Mustafa, Riquelme, Puigcerver, Jenatton, and
  Houlsby}]{DBLP:conf/nips/MustafaRPJH22}
Basil Mustafa, Carlos Riquelme, Joan Puigcerver, Rodolphe Jenatton, and Neil
  Houlsby. 2022.
\newblock Multimodal contrastive learning with limoe: the language-image
  mixture of experts.
\newblock In \emph{NeurIPS}.

\bibitem[{Papineni et~al.(2002)Papineni, Roukos, Ward, and
  Zhu}]{DBLP:conf/acl/PapineniRWZ02}
Kishore Papineni, Salim Roukos, Todd Ward, and Wei{-}Jing Zhu. 2002.
\newblock Bleu: a method for automatic evaluation of machine translation.
\newblock In \emph{{ACL}}, pages 311--318. {ACL}.

\bibitem[{Peebles and Xie(2023)}]{DBLP:conf/iccv/PeeblesX23}
William Peebles and Saining Xie. 2023.
\newblock Scalable diffusion models with transformers.
\newblock In \emph{{ICCV}}, pages 4172--4182. {IEEE}.

\bibitem[{Peng et~al.(2023)Peng, Wang, Dong, Hao, Huang, Ma, and
  Wei}]{DBLP:journals/corr/abs-2306-14824}
Zhiliang Peng, Wenhui Wang, Li~Dong, Yaru Hao, Shaohan Huang, Shuming Ma, and
  Furu Wei. 2023.
\newblock Kosmos-2: Grounding multimodal large language models to the world.
\newblock \emph{CoRR}, abs/2306.14824.

\bibitem[{Qi et~al.(2017)Qi, Yi, Su, and Guibas}]{DBLP:conf/nips/QiYSG17}
Charles~Ruizhongtai Qi, Li~Yi, Hao Su, and Leonidas~J. Guibas. 2017.
\newblock Pointnet++: Deep hierarchical feature learning on point sets in a
  metric space.
\newblock In \emph{{NIPS}}, pages 5099--5108.

\bibitem[{Radford et~al.(2021)Radford, Kim, Hallacy, Ramesh, Goh, Agarwal,
  Sastry, Askell, Mishkin, Clark, Krueger, and
  Sutskever}]{DBLP:conf/icml/RadfordKHRGASAM21}
Alec Radford, Jong~Wook Kim, Chris Hallacy, Aditya Ramesh, Gabriel Goh,
  Sandhini Agarwal, Girish Sastry, Amanda Askell, Pamela Mishkin, Jack Clark,
  Gretchen Krueger, and Ilya Sutskever. 2021.
\newblock Learning transferable visual models from natural language
  supervision.
\newblock In \emph{{ICML}}, volume 139 of \emph{Proceedings of Machine Learning
  Research}, pages 8748--8763. {PMLR}.

\bibitem[{Rajbhandari et~al.(2020)Rajbhandari, Rasley, Ruwase, and
  He}]{DBLP:conf/sc/RajbhandariRRH20}
Samyam Rajbhandari, Jeff Rasley, Olatunji Ruwase, and Yuxiong He. 2020.
\newblock Zero: memory optimizations toward training trillion parameter models.
\newblock In \emph{{SC}}, page~20. {IEEE/ACM}.

\bibitem[{Reuss et~al.(2024)Reuss, Yagmurlu, Wenzel, and
  Lioutikov}]{DBLP:journals/corr/abs-2407-05996}
Moritz Reuss, {\"{O}}mer~Erdin{\c{c}} Yagmurlu, Fabian Wenzel, and Rudolf
  Lioutikov. 2024.
\newblock Multimodal diffusion transformer: Learning versatile behavior from
  multimodal goals.
\newblock \emph{CoRR}, abs/2407.05996.

\bibitem[{Ronneberger et~al.(2015)Ronneberger, Fischer, and
  Brox}]{DBLP:conf/miccai/RonnebergerFB15}
Olaf Ronneberger, Philipp Fischer, and Thomas Brox. 2015.
\newblock U-net: Convolutional networks for biomedical image segmentation.
\newblock In \emph{{MICCAI} {(3)}}, volume 9351 of \emph{Lecture Notes in
  Computer Science}, pages 234--241. Springer.

\bibitem[{Shen et~al.(2023)Shen, Hou, Zhou, Du, Longpre, Wei, Chung, Zoph,
  Fedus, Chen, Vu, Wu, Chen, Webson, Li, Zhao, Yu, Keutzer, Darrell, and
  Zhou}]{DBLP:journals/corr/abs-2305-14705}
Sheng Shen, Le~Hou, Yanqi Zhou, Nan Du, Shayne Longpre, Jason Wei, Hyung~Won
  Chung, Barret Zoph, William Fedus, Xinyun Chen, Tu~Vu, Yuexin Wu, Wuyang
  Chen, Albert Webson, Yunxuan Li, Vincent~Y. Zhao, Hongkun Yu, Kurt Keutzer,
  Trevor Darrell, and Denny Zhou. 2023.
\newblock Flan-moe: Scaling instruction-finetuned language models with sparse
  mixture of experts.
\newblock \emph{CoRR}, abs/2305.14705.

\bibitem[{Skean et~al.(2025)Skean, Arefin, Zhao, Patel, Naghiyev, LeCun, and
  Shwartz{-}Ziv}]{DBLP:journals/corr/abs-2502-02013}
Oscar Skean, Md~Rifat Arefin, Dan Zhao, Niket Patel, Jalal Naghiyev, Yann
  LeCun, and Ravid Shwartz{-}Ziv. 2025.
\newblock Layer by layer: Uncovering hidden representations in language models.
\newblock \emph{CoRR}, abs/2502.02013.

\bibitem[{Song et~al.(2021)Song, Meng, and Ermon}]{DBLP:conf/iclr/SongME21}
Jiaming Song, Chenlin Meng, and Stefano Ermon. 2021.
\newblock Denoising diffusion implicit models.
\newblock In \emph{{ICLR}}. OpenReview.net.

\bibitem[{Su et~al.(2023)Su, Lan, Li, Xu, Wang, and
  Cai}]{DBLP:journals/corr/abs-2305-16355}
Yixuan Su, Tian Lan, Huayang Li, Jialu Xu, Yan Wang, and Deng Cai. 2023.
\newblock Pandagpt: One model to instruction-follow them all.
\newblock \emph{CoRR}, abs/2305.16355.

\bibitem[{Touvron et~al.(2023)Touvron, Lavril, Izacard, Martinet, Lachaux,
  Lacroix, Rozi{\`{e}}re, Goyal, Hambro, Azhar, Rodriguez, Joulin, Grave, and
  Lample}]{DBLP:journals/corr/abs-2302-13971}
Hugo Touvron, Thibaut Lavril, Gautier Izacard, Xavier Martinet, Marie{-}Anne
  Lachaux, Timoth{\'{e}}e Lacroix, Baptiste Rozi{\`{e}}re, Naman Goyal, Eric
  Hambro, Faisal Azhar, Aur{\'{e}}lien Rodriguez, Armand Joulin, Edouard Grave,
  and Guillaume Lample. 2023.
\newblock Llama: Open and efficient foundation language models.
\newblock \emph{CoRR}, abs/2302.13971.

\bibitem[{Vaswani et~al.(2017)Vaswani, Shazeer, Parmar, Uszkoreit, Jones,
  Gomez, Kaiser, and Polosukhin}]{DBLP:conf/nips/VaswaniSPUJGKP17}
Ashish Vaswani, Noam Shazeer, Niki Parmar, Jakob Uszkoreit, Llion Jones,
  Aidan~N. Gomez, Lukasz Kaiser, and Illia Polosukhin. 2017.
\newblock Attention is all you need.
\newblock In \emph{{NIPS}}, pages 5998--6008.

\bibitem[{Vedantam et~al.(2015)Vedantam, Zitnick, and
  Parikh}]{DBLP:conf/cvpr/VedantamZP15}
Ramakrishna Vedantam, C.~Lawrence Zitnick, and Devi Parikh. 2015.
\newblock Cider: Consensus-based image description evaluation.
\newblock In \emph{{CVPR}}, pages 4566--4575. {IEEE} Computer Society.

\bibitem[{Wald et~al.(2019)Wald, Avetisyan, Navab, Tombari, and
  Nie{\ss}ner}]{DBLP:conf/iccv/WaldANTN19}
Johanna Wald, Armen Avetisyan, Nassir Navab, Federico Tombari, and Matthias
  Nie{\ss}ner. 2019.
\newblock {RIO:} 3d object instance re-localization in changing indoor
  environments.
\newblock In \emph{{ICCV}}, pages 7657--7666. {IEEE}.

\bibitem[{Wang et~al.(2022)Wang, Ma, Huang, Dong, Wang, Peng, Wu, Bajaj,
  Singhal, Benhaim, Patra, Liu, Chaudhary, Song, and
  Wei}]{DBLP:journals/corr/abs-2210-06423}
Hongyu Wang, Shuming Ma, Shaohan Huang, Li~Dong, Wenhui Wang, Zhiliang Peng,
  Yu~Wu, Payal Bajaj, Saksham Singhal, Alon Benhaim, Barun Patra, Zhun Liu,
  Vishrav Chaudhary, Xia Song, and Furu Wei. 2022.
\newblock Foundation transformers.
\newblock \emph{CoRR}, abs/2210.06423.

\bibitem[{Wang et~al.(2023)Wang, Bao, Dong, Bjorck, Peng, Liu, Aggarwal,
  Mohammed, Singhal, Som, and Wei}]{DBLP:conf/cvpr/WangBDBPLAMSSW23}
Wenhui Wang, Hangbo Bao, Li~Dong, Johan Bjorck, Zhiliang Peng, Qiang Liu, Kriti
  Aggarwal, Owais~Khan Mohammed, Saksham Singhal, Subhojit Som, and Furu Wei.
  2023.
\newblock Image as a foreign language: {BEIT} pretraining for vision and
  vision-language tasks.
\newblock In \emph{{CVPR}}, pages 19175--19186. {IEEE}.

\bibitem[{Wang et~al.(2025)Wang, Li, Xu, Qi, Yang, Ma, Liu, and
  Zhang}]{DBLP:conf/icmcs/WangLXQYMLZ25}
Xiaoyan Wang, Zeju Li, Yifan Xu, Jiaxing Qi, Zhifei Yang, Ruifei Ma, Xiangde
  Liu, and Chao Zhang. 2025.
\newblock Spatial 3d-llm: Exploring spatial awareness in 3d vision-language
  models.
\newblock In \emph{{ICME}}, pages 1--6. {IEEE}.

\bibitem[{Wu et~al.(2023)Wu, Yin, Qi, Wang, Tang, and
  Duan}]{DBLP:journals/corr/abs-2303-04671}
Chenfei Wu, Shengming Yin, Weizhen Qi, Xiaodong Wang, Zecheng Tang, and Nan
  Duan. 2023.
\newblock Visual chatgpt: Talking, drawing and editing with visual foundation
  models.
\newblock \emph{CoRR}, abs/2303.04671.

\bibitem[{Wu et~al.(2024)Wu, Chen, Pan, Liu, Liu, Dai, Gao, Ma, Wu, Wang, Xie,
  Wu, Hu, Wang, Sun, Li, Piao, Guan, Liu, Xie, You, Dong, Yu, Zhang, Zhao,
  Wang, and Ruan}]{DBLP:journals/corr/abs-2412-10302}
Zhiyu Wu, Xiaokang Chen, Zizheng Pan, Xingchao Liu, Wen Liu, Damai Dai, Huazuo
  Gao, Yiyang Ma, Chengyue Wu, Bingxuan Wang, Zhenda Xie, Yu~Wu, Kai Hu, Jiawei
  Wang, Yaofeng Sun, Yukun Li, Yishi Piao, Kang Guan, Aixin Liu, and 8 others.
  2024.
\newblock Deepseek-vl2: Mixture-of-experts vision-language models for advanced
  multimodal understanding.
\newblock \emph{CoRR}, abs/2412.10302.

\bibitem[{Xue et~al.(2024)Xue, Zheng, Fu, Ni, Zheng, Zhou, and
  You}]{DBLP:conf/icml/XueZFNZZ024}
Fuzhao Xue, Zian Zheng, Yao Fu, Jinjie Ni, Zangwei Zheng, Wangchunshu Zhou, and
  Yang You. 2024.
\newblock Openmoe: An early effort on open mixture-of-experts language models.
\newblock In \emph{{ICML}}. OpenReview.net.

\bibitem[{Yang et~al.(2024{\natexlab{a}})Yang, Yang, Zhang, Hui, Zheng, Yu, Li,
  Liu, Huang, Wei, Lin, Yang, Tu, Zhang, Yang, Yang, Zhou, Lin, Dang, Lu, Bao,
  Yang, Yu, Li, Xue, Zhang, Zhu, Men, Lin, Li, Xia, Ren, Ren, Fan, Su, Zhang,
  Wan, Liu, Cui, Zhang, and Qiu}]{DBLP:journals/corr/abs-2412-15115}
An~Yang, Baosong Yang, Beichen Zhang, Binyuan Hui, Bo~Zheng, Bowen Yu,
  Chengyuan Li, Dayiheng Liu, Fei Huang, Haoran Wei, Huan Lin, Jian Yang,
  Jianhong Tu, Jianwei Zhang, Jianxin Yang, Jiaxi Yang, Jingren Zhou, Junyang
  Lin, Kai Dang, and 22 others. 2024{\natexlab{a}}.
\newblock Qwen2.5 technical report.
\newblock \emph{CoRR}, abs/2412.15115.

\bibitem[{Yang et~al.(2024{\natexlab{b}})Yang, Yang, Gupta, Han, Fei{-}Fei, and
  Xie}]{DBLP:journals/corr/abs-2412-14171}
Jihan Yang, Shusheng Yang, Anjali~W. Gupta, Rilyn Han, Li~Fei{-}Fei, and
  Saining Xie. 2024{\natexlab{b}}.
\newblock Thinking in space: How multimodal large language models see,
  remember, and recall spaces.
\newblock \emph{CoRR}, abs/2412.14171.

\bibitem[{Ye et~al.(2023{\natexlab{a}})Ye, Xu, Xu, Ye, Yan, Zhou, Wang, Hu,
  Shi, Shi, Li, Xu, Chen, Tian, Qi, Zhang, and
  Huang}]{DBLP:journals/corr/abs-2304-14178}
Qinghao Ye, Haiyang Xu, Guohai Xu, Jiabo Ye, Ming Yan, Yiyang Zhou, Junyang
  Wang, Anwen Hu, Pengcheng Shi, Yaya Shi, Chenliang Li, Yuanhong Xu, Hehong
  Chen, Junfeng Tian, Qian Qi, Ji~Zhang, and Fei Huang. 2023{\natexlab{a}}.
\newblock mplug-owl: Modularization empowers large language models with
  multimodality.
\newblock \emph{CoRR}, abs/2304.14178.

\bibitem[{Ye et~al.(2023{\natexlab{b}})Ye, Xu, Ye, Yan, Hu, Liu, Qian, Zhang,
  Huang, and Zhou}]{DBLP:journals/corr/abs-2311-04257}
Qinghao Ye, Haiyang Xu, Jiabo Ye, Ming Yan, Anwen Hu, Haowei Liu, Qi~Qian,
  Ji~Zhang, Fei Huang, and Jingren Zhou. 2023{\natexlab{b}}.
\newblock mplug-owl2: Revolutionizing multi-modal large language model with
  modality collaboration.
\newblock \emph{CoRR}, abs/2311.04257.

\bibitem[{Ze et~al.(2024)Ze, Zhang, Zhang, Hu, Wang, and Xu}]{Ze2024DP3}
Yanjie Ze, Gu~Zhang, Kangning Zhang, Chenyuan Hu, Muhan Wang, and Huazhe Xu.
  2024.
\newblock 3d diffusion policy: Generalizable visuomotor policy learning via
  simple 3d representations.
\newblock In \emph{Proceedings of Robotics: Science and Systems (RSS)}.

\bibitem[{Zhang et~al.(2023{\natexlab{a}})Zhang, Li, and
  Bing}]{DBLP:journals/corr/abs-2306-02858}
Hang Zhang, Xin Li, and Lidong Bing. 2023{\natexlab{a}}.
\newblock Video-llama: An instruction-tuned audio-visual language model for
  video understanding.
\newblock \emph{CoRR}, abs/2306.02858.

\bibitem[{Zhang et~al.(2023{\natexlab{b}})Zhang, Han, Zhou, Hu, Yan, Lu, Li,
  Gao, and Qiao}]{DBLP:journals/corr/abs-2303-16199}
Renrui Zhang, Jiaming Han, Aojun Zhou, Xiangfei Hu, Shilin Yan, Pan Lu,
  Hongsheng Li, Peng Gao, and Yu~Qiao. 2023{\natexlab{b}}.
\newblock Llama-adapter: Efficient fine-tuning of language models with
  zero-init attention.
\newblock \emph{CoRR}, abs/2303.16199.

\bibitem[{Zhang et~al.(2023{\natexlab{c}})Zhang, Wicke, Senel, Figueredo,
  Naceri, Haddadin, Plank, and
  Sch{\"{u}}tze}]{DBLP:journals/corr/abs-2310-12020}
Shengqiang Zhang, Philipp Wicke, L{\"{u}}tfi~Kerem Senel, Luis F.~C. Figueredo,
  Abdeldjallil Naceri, Sami Haddadin, Barbara Plank, and Hinrich Sch{\"{u}}tze.
  2023{\natexlab{c}}.
\newblock Lohoravens: {A} long-horizon language-conditioned benchmark for
  robotic tabletop manipulation.
\newblock \emph{CoRR}, abs/2310.12020.

\bibitem[{Zhao et~al.(2023)Zhao, Yu, Ge, Yang, Wei, Zhou, Sun, Peng, Dong, Han,
  and Zhang}]{DBLP:journals/corr/abs-2307-09474}
Liang Zhao, En~Yu, Zheng Ge, Jinrong Yang, Haoran Wei, Hongyu Zhou, Jianjian
  Sun, Yuang Peng, Runpei Dong, Chunrui Han, and Xiangyu Zhang. 2023.
\newblock Chatspot: Bootstrapping multimodal llms via precise referring
  instruction tuning.
\newblock \emph{CoRR}, abs/2307.09474.

\bibitem[{Zheng et~al.(2024{\natexlab{a}})Zheng, Peng, Yang, Shen, Li, Liu,
  Zhou, Li, and You}]{opensora}
Zangwei Zheng, Xiangyu Peng, Tianji Yang, Chenhui Shen, Shenggui Li, Hongxin
  Liu, Yukun Zhou, Tianyi Li, and Yang You. 2024{\natexlab{a}}.
\newblock \href {https://github.com/hpcaitech/Open-Sora} {Open-sora:
  Democratizing efficient video production for all}.

\bibitem[{Zheng et~al.(2024{\natexlab{b}})Zheng, Peng, Yang, Shen, Li, Liu,
  Zhou, Li, and You}]{DBLP:journals/corr/abs-2412-20404}
Zangwei Zheng, Xiangyu Peng, Tianji Yang, Chenhui Shen, Shenggui Li, Hongxin
  Liu, Yukun Zhou, Tianyi Li, and Yang You. 2024{\natexlab{b}}.
\newblock Open-sora: Democratizing efficient video production for all.
\newblock \emph{CoRR}, abs/2412.20404.

\bibitem[{Zhu et~al.(2023{\natexlab{a}})Zhu, Chen, Shen, Li, and
  Elhoseiny}]{DBLP:journals/corr/abs-2304-10592}
Deyao Zhu, Jun Chen, Xiaoqian Shen, Xiang Li, and Mohamed Elhoseiny.
  2023{\natexlab{a}}.
\newblock Minigpt-4: Enhancing vision-language understanding with advanced
  large language models.
\newblock \emph{CoRR}, abs/2304.10592.

\bibitem[{Zhu et~al.(2024)Zhu, Qu, Dong, Ruan, Tong, He, and
  Cheng}]{DBLP:conf/emnlp/0002QDRTH024}
Tong Zhu, Xiaoye Qu, Daize Dong, Jiacheng Ruan, Jingqi Tong, Conghui He, and
  Yu~Cheng. 2024.
\newblock Llama-moe: Building mixture-of-experts from llama with continual
  pre-training.
\newblock In \emph{{EMNLP}}, pages 15913--15923. Association for Computational
  Linguistics.

\bibitem[{Zhu et~al.(2023{\natexlab{b}})Zhu, Ma, Chen, Deng, Huang, and
  Li}]{DBLP:conf/iccv/ZhuMCD0023}
Ziyu Zhu, Xiaojian Ma, Yixin Chen, Zhidong Deng, Siyuan Huang, and Qing Li.
  2023{\natexlab{b}}.
\newblock 3d-vista: Pre-trained transformer for 3d vision and text alignment.
\newblock In \emph{{ICCV}}, pages 2899--2909. {IEEE}.

\end{thebibliography}
\end{document}